\documentclass[letterpaper]{article} 
\usepackage{aaai25}  
\usepackage{times}  
\usepackage{helvet}  
\usepackage{courier}  
\usepackage[hyphens]{url}  
\usepackage{graphicx} 
\usepackage{natbib}  
\usepackage{caption} 
\usepackage{algorithm}
\usepackage{algorithmic}

\usepackage{newfloat}
\usepackage{listings}
\DeclareCaptionStyle{ruled}{labelfont=normalfont,labelsep=colon,strut=off} 
\floatstyle{ruled}
\newfloat{listing}{tb}{lst}{}
\floatname{listing}{Listing}
\usepackage{graphicx}
\usepackage{algorithm, algorithmic}
\usepackage{amsmath}
\usepackage{amssymb}
\usepackage{mathtools}
\usepackage{amsthm}
\usepackage{multirow}
\usepackage{multicol}
\usepackage{amsfonts}
\usepackage{bm}
\usepackage{makecell}
\usepackage{bbding}
\usepackage{color}
\usepackage{thm-restate}
\usepackage{enumerate}
\usepackage{changepage}
\usepackage{varwidth}
\usepackage{booktabs}

\usepackage{xcolor}

\newtheorem{assumption}{Assumption}
\newtheorem{proposition}{Proposition}
\newtheorem{corollary}{Corollary}
\newtheorem{lemma}{Lemma}
\newtheorem{definition}{Definition}
\newtheorem*{remark*}{Remark}

\usepackage{color}

\def\va{{\mathbf{a}}}

\def\sA{{\mathbb{A}}}
\def\sK{{\mathbb{K}}}
\def\sX{{\mathbb{X}}}

\def\sD{{\mathbb{D}}}

\def\sM{{\mathbb{M}}}
\def\sR{\mathbb{R}}
\def\sB{\mathbb{B}}
\def\sH{\mathbb{H}}

\def\sR{\mathbb{R}}

\def\va{{\mathbf{a}}}

\def\vx{{\mathbf{x}}}
\def\vy{{\mathbf{y}}}

\def\mA{{\mathbf{A}}}

\def\sF{{\mathcal{F}}}

\def\eE{\mathop{\mathbb{E}}\limits}

\def\tD{\tilde{\sD}}

\def\tvx{\tilde{\mathbf{x}}}
\def\ty{\tilde{y}}

\def\T{\mathcal{T}}
\def\M{\mathcal{M}}

\def\eva{{a}}
\def\RM{\mathtt{RM}}
\def\RU{\mathtt{RU}}

\DeclareMathOperator*{\sgn}{sgn}

\def\L{{\mathcal{L}}}

\title{Towards Macro-AUC oriented Imbalanced Multi-Label Continual Learning}
\author {
    Yan Zhang\textsuperscript{\rm 1},
    Guoqiang Wu\textsuperscript{\rm 1}\thanks{Corresponding Author},
    Bingzheng Wang\textsuperscript{\rm 1}
    Teng Pang\textsuperscript{\rm 1}
    Haoliang Sun\textsuperscript{\rm 1}
    Yilong Yin\textsuperscript{\rm 1}\footnotemark[1]
}
\affiliations {
    \textsuperscript{\rm 1}School of Software, Shandong University\\
    yannzhang9@gmail.com, 
    guoqiangwu90@gmail.com, 
    binzhwang@gmail.com,
    silencept7@gmail.com,\\
    haolsun@sdu.edu.cn,
    ylyin@sdu.edu.cn

}

\usepackage{bibentry}

\begin{document}

\maketitle

\begin{abstract}
    In Continual Learning (CL), while existing work primarily focuses on the multi-class classification task, there has been limited research on Multi-Label Learning (MLL). In practice, MLL datasets are often class-imbalanced, making it inherently challenging, a problem that is even more acute in CL. 
    Due to its sensitivity to imbalance, Macro-AUC is an appropriate and widely used measure in MLL. 
    However, there is no research to optimize Macro-AUC in MLCL specifically.
    To fill this gap, in this paper, we propose a new memory replay-based method to tackle the imbalance issue for Macro-AUC-oriented MLCL.
    Specifically, inspired by recent theory work, we propose a new Reweighted Label-Distribution-Aware Margin (RLDAM) loss.
    Furthermore, to be compatible with the RLDAM loss, a new memory-updating strategy named Weight Retain Updating (WRU) is proposed to maintain the numbers of positive and negative instances of the original dataset in memory. 
    Theoretically, we provide superior generalization analyses of the RLDAM-based algorithm in terms of Macro-AUC, separately in batch MLL and MLCL settings. This is the first work to offer theoretical generalization analyses in MLCL to our knowledge.
    Finally, a series of experimental results illustrate the effectiveness of our method over several baselines.
    Our codes are available at \textit{https://github.com/ML-Group-SDU/Macro-AUC-CL}.
\end{abstract}

%

\section{Introduction}
Traditional machine learning methods assume an independent and identically distributed (i.i.d.) data pattern.
However, in practice, humans continually learn new knowledge and retain previous knowledge.
This process, known as Continual Learning (CL)~\citep{ring1994cl}, aims to adapt to new tasks while mitigating Catastrophic Forgetting~\citep{french1999catastrophic}.

Recently, CL has gained considerable attention~\citep{wang2023comprehensive}, where the replay-based approaches~\citep{chaudhry2019ER,rebuffi2017icarl} often show promising performance~\citep{knoblauch2020clhard}.
Most of these works concentrate on multi-class classification, where an instance is associated with a single label. 
In some real-world scenarios, however, an instance is often associated with multiple labels simultaneously. This raises the study of Multi-Label Learning (MLL)~\citep{mccallum1999mll}. In this paper, our focus lies on \emph{Multi-Label Continual Learning (MLCL)}~\citep{kim2020prs}, an inherently more challenging learning task that has not yet been extensively explored. 
MLCL is increasingly relevant in real-world applications. This setting is common in areas such as healthcare, finance, and autonomous systems, where the ability to handle multiple labels as new labels emerge is crucial \citep{kassim2024multiappli1,dalle2023multiappli2,ceccon2024multiappli3}.

In MLCL (or MLL), there is a natural and inherent challenge - class imbalance, which is a special characteristic of MLL, unlike the imbalance case of multi-class problem. 
The imbalance issue in MLCL we considered is for each MLL task in the continual setting, which mainly includes two aspects: imbalance within labels, and imbalance between labels. 
Specifically, imbalance within labels is that for each label, the numbers of negative instances are often higher than the positive instances\footnote{Positive and negative instances refer to samples that either contain or lack the corresponding label.}. In contrast, imbalance between labels is that the number of positive instances of each label is not balanced.
For the performance evaluation on imbalanced cases, various measures are developed, e.g., F1 score and mAP, etc. 
Among them, Macro-AUC~\citep{zhang2013review} serves as a suitable and widely-used measure in practice, which is considered in this paper. 

Optimizing Macro-AUC in MLL directly can lead to NP-hard problems~\citep{tarekegn2021review}, as it is discontinuous and non-convex. 
So it is common to design a surrogate loss to solve this issue.
Meanwhile, it is crucial to address the imbalance problem to optimize Macro-AUC. However, commonly used losses such as cross-entropy loss are ineffective in handling imbalance, leading to relatively low Macro-AUC performance (see Table~\ref{table1} in Section~\ref{sec:experiments}).
To tackle imbalance and maximize Macro-AUC for \emph{multi-label batch learning}, recent theory work~\citep{wu2023towards} proposes a reweighted univariate (RU) loss with superior formal learning guarantees. 
Both theory and experiments show that RU loss enjoys the high performance of pairwise loss, without its high computational complexity, and the computational efficiency of univariate losses (e.g., cross-entropy), which suffer from lower performance. 
However, this loss assigns the same margin to each class, which may require further analysis to enhance its performance.
In contrast, in \emph{ multi-class batch learning}, a label-distribution-aware margin (LDAM) loss, proposed by~\citet{cao2019marginloss}, assigns a larger margin to the class with a smaller size, resulting in different margins for distinct classes. It has shown promising performance in practice.

Inspired by the above studies, we propose a novel replay-based method that aims to maximize Macro-AUC in MLCL setting.
Our method primarily consists of a new loss function and a new memory-updating strategy.
Specifically, we propose a new reweighted label-distribution-aware margin (RLDAM) loss to tackle the imbalance, which can inherit the benefits of the RU and LDAM losses. 
Further, to ensure better compatibility with the RLDAM loss, we propose a new memory-updating approach named Weight Retain Updating (WRU), which is simple but effective.
Intuitively, it ensures that the data stored in memory for each class maintains consistency with the original dataset in terms of the number of positive and negative instances (a.k.a. reweighting factors). Theoretically, we provide superior generalization analyses of the RLDAM-based algorithm for Macro-AUC in batch MLL setting, and then extend it to MLCL setting.


Finally, to illustrate the effectiveness of our proposed method, we conduct a series of experiments. Comparisons with other baselines demonstrate the superiority of our approach. Moreover, we have performed ablation studies and experiments to investigate other influencing factors, consistently showing that our method performs well.


Our contribution can be summarized as:
(1) To maximize Macro-AUC in MLCL, we propose a novel memory replay-based method, involving a new theory-principled RLDAM loss and a new compatible memory updating strategy called WRU.
(2) Theoretically, we analyze the generalization superiority of the RLDAM-based algorithm w.r.t. Macro-AUC in batch MLL and MLCL. To our knowledge, ours is the first theoretical analysis of MLCL.
(3) Experimentally, extensive results demonstrate the superiority of our method over several baselines in MLCL.

\section{Preliminaries}
\textbf{Notations.}
Let a regular, boldfaced lower and upper letter denote a scaler (e.g., $a$), vector (e.g., $ \va $), and matrix (e.g., $\mA$), respectively.  $\eva_i$ denotes its $i$-th element. For a matrix $\mA$, denote its $i$-th row, $j$-th column, and $(i,j)$-th element as $\va_i,\va^j$, and $a_{ij}$, respectively. Let a blackboard bold letter denote a set (e.g., $\sA$). 
$|\sA|$ denotes its cardinality. $[n]$ denotes the set $\{1, ..., n\}$. $[\![\cdot]\!]$ denotes the indicator function, i.e., it returns 1 if the condition holds and 0 otherwise.

\subsection{Problem Setup}

\textbf{Multi-Label Learning (MLL).}
Given a MLL training dataset $\sD=\{(\vx_i,\vy_i)\}_{i=1}^n$ i.i.d. drawn from $P$ with a sample size $n$, where $P$ is a distribution over $\mathcal{X} \times \{0,1\}^{K}$ and $K$ is the size of label set $\sK$. 
${\vx}_i$ and $\vy_i$ denote the input and corresponding output. Note that $\vy_i \in \{0,1\}^{K}$ is a multi-hot label vector.
For any $k \in \sK$, $y_{ik} = 1$ (or $0$) denotes that the $k$-th label is relevant (or irrelevant).
The objective of MLL is to learn a hypothesis or predictor $f= (f_1,...,f_K):\mathcal{X} \to \sR^K$ from a hypothesis space $\sF:= \{ f \}$.
In batch MLL, the learner knows the label size $K$ in advance.

\noindent\textbf{Multi-Label Continual Learning (MLCL).}
In MLCL, the learner continuously encounters a sequence of tasks $\{ \T^1,...,\T^T \}$, where $T$ denotes the length of the task sequence, and each task $\T^t$ ($t \in [T]$) is an MLL task. 

Notably, when a task identifier (e.g., $t$) appears in the superscript, it does not mean the power of $t$ but denotes it is for the $t$-th task, and this definition holds for all symbols. Moreover, we place $+$ or $-$ in the superscript position to indicate that the symbol is about the positive or negative instances of a particular label (e.g. $n_k^{t+}$ is the sample size of the positive instances for the class $k$ in the $t$-th task).

Specifically, at each time step $t$, the learner encounters a new task $\T^t$ with a dataset $\sD^t=\{(\vx_i^t,\vy_i^t)\}_{i=1}^{n^t}$, which is i.i.d. drawn from a MLL distribution $P^t$ over $\mathcal{X} \times \{0,1\}^{K^t}$, where $K^t$ is the label size for the current task $\T^t$.
The goal of MLCL is to learn a hypothesis $f^t=(f_1^t,...,f_{\widetilde{K}^t}^t): \mathcal{X} \to \sR^{\widetilde{{K}}^t}$ over all encountered tasks from a hypothesis space $\sF^t := \{ f^t \}$, where $\widetilde{K}^t = \sum_{i=1}^{t} K^i$.
Notably, the learning objective reflects that the hypothesis applies to both the new task and all previously encountered tasks, even if only $\sD^t$ is accessible currently.

\noindent\textbf{Multi-Label Class Incremental Learning (MLCIL).} 
Among MLCL, here we focus on one practical and challenging case - MLCIL
where the prediction layer of the learner is a \emph{single-head} setup, and the task ID is not accessible during the learning process.
Similarly to MLL, MLCIL assumes that the continual learner knows the total number of classes $K = \sum_{t \in [T]} K^t$ in advance.
However, unlike batch MLL, the learner of MLCIL learns only task-known classes in each task and does not consider task-agnostic classes.
Besides the above description, we also make the following assumptions.
\begin{assumption}[\textbf{For MLCIL}]
\label{assump_repetition}
    \begin{enumerate}[(1)]
    \setlength\itemsep{-2pt}
        \item 
        \label{assump_re_1}
        Each task has distinct classes: $\forall t \ne t^{\prime},\sK^t \cap \sK^{t^{\prime}}=\emptyset$.
        \item 
        \label{assump_re_2}
        Within each task, there exists at least one sample associated with more than one label (multi-label setup)~\footnote{This assumption is reasonable as most practical MLL problems hold. While it is theoretically possible to sample datasets where the number of labels per sample is less than two in a probabilistic fashion, this case is extremely rare in practice.}: $\forall t \in [T], \exists (\vx^t_i,\vy^t_i) \in \sD^t,\sum_j y^t_{ij} > 1$. 
        \item 
        \label{assump_re_3}
        Samples may appear repeatedly across different tasks (same as \citet{dong2023krt}): $\exists t \ne t^{\prime}, \ \sX^{t} \cap \sX^{t^{\prime}} \ne \emptyset$. 
    \end{enumerate}
\end{assumption}
Assumption (\ref{assump_re_2}) and (\ref{assump_re_3}) reflects the differences between MLCIL and multi-class CL.
In multi-class CL settings, an instance exhibits only one label and task-disjoint assumption often holds, where both are opposite to MLCIL.

\noindent\textbf{Replay-Based CL.} We consider the replay-based CL framework, specifically rehearsal, because it is simple and effective. It maintains a memory buffer $\M$ for selected samples of all previous tasks.
The data of each task $\T^t$ saved in memory is denoted as $\sM^t$, which is a sampled subset from $\sD^t$.

\subsection{Evaluation Measure}
Various measures are often used in MLCIL for a comprehensive evaluation, e.g., F1 score, and mAP. 
Concerning the (label-wise) class imbalance, Macro-AUC~\citep{zhang2013review} is an appropriate and widely used measure in practice but has not yet been considered in MLCIL, which we focus on. 
Macro-AUC macro-averages the AUC measure across all class labels. Given a dataset $\sD$ and a predictor $f \in \sF$, Macro-AUC is defined as follows:
\begin{equation}
    \frac{1}{K} \sum_{k=1}^{K} \frac{1}{\left|\sD_{k}^{+}\right|\left|\sD_{k}^{-}\right|} \sum_{(p, q) \in \sD_{k}^{+} \times \sD_{k}^{-}}[\![f_{k}\left({\vx}_{p}\right)>f_{k}\left({\vx}_{q}\right)]\!],
\label{eq:eva}
\nonumber
\end{equation}
where for each class $k$, $\sD_k^+$ and $\sD_k^-$ denote the index subset of positive and negative instances, respectively, and $f_k(\vx)$ is the logit output of input $\vx$. 
The objective for maximizing Macro-AUC is:
$
    \frac{1}{K} \sum_{k=1}^K \frac{1}{|\sD_k^+| |\sD_k^-|} \sum_{(p, q) \in \sD_k^+ \times \sD_k^-} \L_{0/1}(\vx_p, \vx_q, f_k), 
$
where the $0/1$ loss is $\L_{0/1}(\vx^+, \vx^-, f_k) = [\![ f_k(\vx^+) \leq f_k(\vx^-) ]\!].$
The associated expected risk w.r.t. the $0/1$ loss is
\begin{align}
    R^{0/1}(f) = \frac{1}{K} \sum_{k=1}^K \eE_{\vx_p \sim P_k^+, \vx_q \sim P_k^-} \left[ \L_{0/1}(\vx_p, \vx_q, f_k) \right],
\end{align}
where $P_k^+ = P(\vx|y_k=1),P_k^- = P(\vx|y_k=0)$.
When measuring the test performance, we use an overall Macro-AUC, i.e., the mean of Macro-AUC on all tasks.
Besides, we use Forgetting~\citep{chaudhry2018rwalk} to evaluate the memory ability of the continual learning technique.

\section{Method}
In this section, we propose our method to maximize Macro-AUC for MLCIL. Firstly, we propose a new reweighted label-distribution-aware margin (RLDAM) loss to tackle the multi-label imbalance by combining the reweighted loss and LDAM loss. Regarding the loss, a new memory updating strategy named Weight Retain Updating (WRU) is proposed for replay-based MLCIL.

\subsection{Reweighted Label-Distribution-Aware Margin Loss}
For a clear presentation, we first propose a new RLDAM loss in the batch MLL scenario and then discuss it in the continual learning scenario. 

\subsubsection{Batch Learning Scenario}
In the batch learning scenario, to maximize Macro-AUC, recent theory work~\citep{wu2023towards} proposed a Reweighted Univariate (RU) loss with good properties as $\L_{\RU}(\vx^+,\vx^-,f_k)=\ell(f_k(\vx^+))+\ell(-f_k(\vx^-)),$
where $\vx^+$ (or $\vx^-$) denotes a positive (or negative) instance for the label $k$, and the base loss $\ell(\cdot)$ can be any \emph{margin}-based loss (e.g., hinge loss or logistic loss), for binary classification.
The empirical risk w.r.t. $\L_{\RU}$ is 
$
    \widehat{R}_\sD^{\RU}(f)
    =  \frac{1}  {K}\sum_{k=1}^{K}\sum_{i=1}^{n} \bigg( \frac{[[y_{ik}=1]]}{|\sD_k^+|}\ell(f_k(\vx_i)) 
     + \frac{[[y_{ik}\ne 1]]}{|\sD_k^-|}\ell(-f_k(\vx_i)) \bigg)
$
where $1/|\sD_k^+|$ and $1/|\sD_k^-|$ are reweighted factors.
Intriguingly, the theory work~\citep{wu2023towards} shows that the learning algorithm with the RU loss offers a better learning guarantee than the one with the original univariate loss. However, for the base loss of RU considered in~\citet{wu2023towards}, the margin for each class is the same, which might need further analysis for improvement.


On the other hand, to tackle the imbalance issue, a label-distribution-aware margin (LDAM) loss is proposed by~\citet{cao2019marginloss}, where their theory suggests that it can improve generalization by assigning different margins for distinct classes.
In that work, for a loss function in binary classification, such as hinge loss $\L_{\mathtt{HG}}(\vx,\vy,f)=\max(0, \Delta - yf(\vx))$, the optimal trade-off of two margins is derived as $\Delta_1=\frac{\lambda}{n_1^{1/4}}$ and $\Delta_2=\frac{\lambda}{n_2^{1/4}}$ with a constant $\lambda$ for distinct classes, where $n_1$ and $n_2$ are the number of samples for two classes. 
Then it extends the margins to multi-class hinge loss as: 
\begin{equation}
\L_\mathtt{LDAM-HG}(\vx,\vy,f)=\max({\mathop{\max}\limits_{j\ne k}} \{f_j\}-f_k+\Delta_k,0),
\label{eq:margin}
\end{equation}
where $\forall k \in \sK$, $\Delta_k=\frac{\lambda}{n_k^{1/4}}$ and $n_k$ denotes the number of instances for class $k$. 
In the following of our work, we take inspiration from the optimal trade-off between two classes and its multi-class extension.

\textbf{Our new RLDAM loss.} To combine the best of the above two losses, we integrate LDAM loss into RU loss to further improve generalization for the first time.
Applying LDAM to the base loss of RU loss, we obtain a new Reweighted Label-Distribution-Aware Margin Loss (RLDAM) as 
\begin{align}
    \nonumber \L_{\RM}\left({\vx}^{+}, {\vx}^{-}, f_{k}\right) 
    = & \ell\left(f_{k}\left({\vx}^{+}\right)-\Delta_k^+\right) + \\
    & \ell\left(-f_{k}\left({\vx}^{-}\right)-\Delta_k^-\right),
    \label{eq:L_rm}
\end{align}
where $\Delta_k^+$ and $\Delta_k^-$ are the margin of positive and negative instances of class $k$. 
Note that different classes exhibit different margins.~\footnote{We will derive the optimal value of $\Delta_k^+$ and $\Delta_k^-$ and prove that minimizing the RLADM loss can have a better learning guarantee w.r.t. Macro-AUC than that of the RU loss in Section~\ref{sec:theoretical_analyses}.}

Next, we will analyze the RLDAM loss in the continual learning setting.
Before moving to the continual learning scenario, we first consider the scenario where multiple tasks are learned jointly in the batch learning mode. And the empirical risk of each class $k$ in a task $\T^t$ is 
\begin{align}
\nonumber
\widehat{R}_{{\sD}_k^t}^{\RM}(f) = 
& \frac{1}{\left| \sD_{k}^{t+} \right| \left| \sD_{k}^{t-} \right|} \sum_{(p, q) \in \sD_{k}^{t+} \times \sD_{k}^{t-} } \mathcal{L}_{\RM}\left({\vx}_{p}, {\vx}_{q}, f_{k}\right) \\ \nonumber
= &  \sum_{i=1}^{n^t}\bigg([[y_{ik}=1]] \frac{1}{| {\sD_{k}^{t+}} |} \ell \left(f_{k}\left({\vx}_{i}\right)-\Delta_k^{t+}\right)+ \\ 
& [[y_{ik} \neq 1]] \frac{1}{| {\sD_{k}^{t-}} |} \ell \left(-f_{k}({\vx}_{i})-\Delta_k^{t-}\right) \bigg),
\label{eq:emp_ldam}
\end{align}
where $\sD^t_k$ is the subset of the dataset for class $k$ in the $t$-th task $\T^t$.
Then, the empirical risk of each task $\T^t$ and overall risk for $T$ tasks are:
\begin{align}
    & \widehat{R}_{{\sD}^t}^{\RM}(f) = \frac{1}{|\sK^t|}\sum_{k\in{\sK^t}}\widehat{R}_{\sD^t_k}^{\RM}(f), \\
    & \widehat{R}_{{\T}}^{\RM}(f) = \frac{1}{T} \left( \widehat{R}_{{\sD}^T}^{\RM}(f) +  \sum_{i=1}^{T-1} \widehat{R}_{{\sD}^i}^{\RM}(f) \right).
    \label{eq:losst}
\end{align}

\subsubsection{Continual Learning Scenario}
For a general CL scenario, at each time step $t$, the empirical risk $\widehat{R}_{\T^t}^{\RM}$ originally is equal to $\widehat{R}_{\sD^t}^{\RM}(f)$ in Eq.~\eqref{eq:losst} as only the data in $\T^t$ can be accessed.
Here we focus on the memory replay-based framework, which explicitly stores a subset $\sM^i$ of $\sD^i$ for each previous task $\T^i \ (i \in [t-1])$. The (adjusted) empirical risk can be written as $\widehat{R}_{{\T^t}}^{\RM}(f) = \frac{1}{t} \left( \widehat{R}_{\sD^t}^{\RM}(f) +  \sum_{i=1}^{t-1} \widehat{R}_{{\sM}^i}^{\RM}(f) \right).$

Note that we should make the risk on subset $\sM^i$ serving as an (approximate) unbiased estimator of the risk on whole dataset $\sD^i$, i.e., $\eE_{\sM^i} \left[ \widehat{R}_{\sM^i}^{\RM}(f) \right] \approx \widehat{R}_{\sD^i}^{\RM}(f)$. Since past $\sD^i$ is inaccessible in CL, we instead minimize the empirical risk on $\sM^i$, 
which can also reduce the risk on $\sD^i$ with high probability. 
We will discuss it in detail in the subsequent section.

\subsection{Weight Retain Updating for Memory}
\label{sec:wru}

To achieve the goal of $\eE_{\sM^i} \left[ \widehat{R}_{\sM^i}^{\RM}(f) \right] \approx \widehat{R}_{\sD^i}^{\RM}(f)$, we should maintain consistency in the $|\sD_k^+|,|\sD_k^-|$ of Eq.~\eqref{eq:emp_ldam}, which is the critical reweighting factor, between $\sM_k$ and $\sD_k$ for each class $k \in \sK^t$. Note that the widely-used strategies for updating memory buffer in CL, such as Reservoir Sampling (RS)~\citep{vitter1985reservoir}, do not consider this, resulting in varying degrees of negative impact on the RLDAM loss of data stored in memory. Therefore, we propose a new Weight Retain Updating (WRU) strategy to address this issue.

Specifically, after learning a task, for each class $k \in \sK^t$, we first calculate the $|\sD_k^+|$ and  $|\sD_k^-|$, respectively.
Then, we design a greedy algorithm to select $|{\sM}^t|$ samples from $\sD^t$. The goal of the greedy algorithm is to minimize the discrepancy in the ratio of positive and negative instances between $\sD^t$ and ${\sM}^t$. The selection principle is outlined as $s^*=\mathop{\arg\min}\limits_{s} \sum_{k\in\sK^t} \left| \mathtt{Rat}(\sD^t,k)-\mathtt{Rat}({\sM}^t\cup (\vx_s,\vy_s),k) \right|,$
where $s^*$ is the index of our best choice in each selection step, and the function $\mathtt{Rat}(\sD,k) = {|\sD_k^+|}/{|\sD_k^-|},\sD_k^+ \subset \sX,\sD_k^- \subset \sX$ takes a dataset and a class index as input, and outputs the ratio. We repeatedly perform this selection for storage until the memory is full.

The above procedure will give an approximate ratio, however, given that the cost of storing a few constants is much lower than storing samples, we rather explicitly store $|\sD_k^+|$ and  $|\sD_k^-|$ into memory along with $|{\sM}^t|$ samples selected using our designed algorithm from $\sD^t$.
Thus we can maintain the original constants of positive and negative instances between $\sD^t$ for its corresponding subset ${\sM}^t$. 

Further, to fully utilize the memory space, we follow \citet{rebuffi2017icarl} to store $M/t$ samples for each task after learning $\T^t$. 
Due to the fixed memory size, we remove some past samples before storing new ones, where we just remove samples without changing the stored $|\sD_k^+|$ and  $|\sD_k^-|$.

Overall, the training procedure of our continual learning method is summarized in Algorithm \ref{alg:algorithm0}.

\begin{algorithm}[tb]
\caption{Replay-based Continual Learning Procedure.}
\label{alg:algorithm0}
\textbf{Input}: Tasks $\mathcal{T}$, Task length $T(T>1)$, Memory $\mathcal{M}$, Memory size $M$;\\
\textbf{Parameter}: Learning rate $\eta$, Batch size $B$, Epochs $n_{e}$, The model $f$;\\
\textbf{Output}: Learned parameter $\Theta$ of $f$;

\begin{algorithmic}[1]  
\STATE \textbf{for} $t \in (1,T)$ \textbf{do}
\STATE $\quad$ Get dataset $\sD^t$ from $\T^t$;
\STATE $\quad$ \textbf{if} $t = 1$ \textbf{then}
\STATE $\quad \quad$ Perform Batch Learning on the $\sD^t$;
\STATE $\quad$ \textbf{else} // Continual learning procedure;
\STATE $\quad \quad$ Get $\{\sM^1,...,\sM^{t-1}\}$ from $\M$;
\STATE $\quad \quad$ // Training iteration
\STATE $\quad \quad$ \textbf{for} $i \in (1, n_{e})$ \textbf{do}
\STATE $\quad \quad \quad$ \textbf{for} each batch $\sB^t \in \sD^t \ (|\sB^t|=B)$ \textbf{do}
\STATE $\qquad \qquad$ Sample a batch $\sB^{\M}$ with size of $B$ from 

$\qquad \qquad$ $\cup_{i \in (1,t-1)}\sM^i$;
\STATE $\qquad \qquad$ Update model parameters $\Theta$ according to 

$\qquad \qquad$ Eq.~(19) (Appendix~E.2) with $\sB^t,\sB^{\M},\eta$;
\STATE $\quad$// Training ending;
\STATE $\quad$Updating the memory according to Sec. \ref{sec:wru}; 
\STATE \textbf{return} $\Theta$;
\end{algorithmic}
\end{algorithm}

\section{Theoretical Analyses}
\label{sec:theoretical_analyses}
Here, we first analyze the generalization bound of the RLDAM-based algorithm in the batch MLL, and then give the generalization of our algorithm with RLDAM loss and WRU in MLCL.
Notably, for batch MLL, technically, compared with the work~\citep{wu2023towards} we mainly follow, 
we introduce a new definition of fractional Rademacher complexity for the hypothesis space with its upper bound, and a new contraction inequality (see Appendix~B for details). 
Moreover, to the best of our knowledge, this is the first theoretical work on MLCL.


\noindent\textbf{For the RLDAM-based algorithm in batch MLL.} Firstly, we introduce the quantity of label-wise class imbalance \citep{wu2023towards} as $\tau_k = \min\{|\sD_k^+|, |\sD_k^-|\} / {n}, \forall k \in [K]$.
In continual learning setting, we denote $\tau_k^i$ as the imbalance of the $i$-th task.
For simplicity, here we consider the general kernel-based hypothesis class, formally written as:
\begin{align}
    \label{eq:kernel_hypothesis}
    \nonumber \mathcal{F} = \big\{ \mathbf{x} \mapsto \mathbf{W} ^\top \Phi(\mathbf{x}): \mathbf{W} = (\mathbf{w}_1, \ldots ,\mathbf{w}_K)^\top,& \\
    \| \mathbf{w}_k \|_{\mathbb{H}} \leq \Lambda & \big\},
\end{align}
where $\kappa:\mathcal{X} \times \mathcal{X} \to \sR$ is a positive definite symmetric kernel and its induced reproducing kernel Hilbert space (RKHS) is $\sH$, and $\Phi:\mathcal{X} \to \sH$ is a feature mapping of $\kappa$.
Note that  although we utilize deep neural networks (DNNs) in experiments, it can still provide valuable insights because recent theory~\citep{jacot2018ntk} has established the connection between over-parameterized DNNs and Neural Tangent Kernel (NTK)-based methods.
More specifically, \citet{huang2020resnetntk1,tirer2022resnetntk2} analyses the ResNet, which is used in our work, from the NTK perspective.
Then, we introduce the following mild assumption for subsequent analyses.
\begin{assumption}
\label{assump_common}
    \begin{enumerate}[(1)]
    \setlength\itemsep{0pt}
        \item The training data $\sD = \{(\mathbf{x}_i, \mathbf{y}_i )\}_{i=1}^n$ is i.i.d. sampled from the distribution $P$, where $\exists \ r > 0$, it satisfies $\kappa (\mathbf{x}, \mathbf{x}) \leq r^2$ for all $\mathbf{x} \in \mathcal{X}$.
        \item The hypothesis class is defined in Eq.~\eqref{eq:kernel_hypothesis}.
	  \item The base loss $\ell (z)$ is the hinge loss and bounded by $B$, and $\forall k \in \sK$, $\ell(z-\Delta_k^+)$ and $\ell(z-\Delta_k^-)$ is $\rho_k^+$- and $\rho_k^-$- Lipschitz continuous with $\rho_k^+ = 1/\Delta_k^+,\rho_k^- = 1/\Delta_k^-$. \footnote{Note that, the widely-used hinge and logistic loss are both $1$-Lipschitz continuous and we analyze the hinge loss here for clarity. Similar analyses can be obtained for other margin-based losses.
    }
    \end{enumerate}
\end{assumption}
Besides, the expected risk w.r.t. RLDAM loss can be defined as $R^{\RM} (f) = \eE_{\sD} \left[\widehat{R}_{\sD}^{\RM} (f)\right]$ and we can get $R^{0/1}(f) \leq R^{\RM} (f)$. Then, we can obtain the learning guarantee of the RLDAM-based algorithm w.r.t. the Macro-AUC measure, informally as follows (see Appendix~A for formal details):
\begin{align*}
    R^{0/1}(f) \leq \widehat{R}_{\sD}^{\RM} (f) + O \left( \frac{1 
    }{\sqrt{n}} \bigg( \frac{1}{K} \sum_{k=1}^{K} \sqrt{\frac{1}{\tau_k}}(\rho_k^+ + \rho_k^-) \bigg) \right).
\end{align*}
This bound indicates that the batch algorithm with $\L_{\RM}$ has an imbalance-margin-aware learning guarantee of $O \left( \left( \frac{1}{K} \sum_{k=1}^K \sqrt{\frac{1}{\tau_k}}(\rho_k^+ + \rho_k^-) \right) \right)$ w.r.t. Macro-AUC. 
Compared with the one of the $\L_{\RU}$-based algorithm (Theorem~4 in \citet{wu2023towards}), with identical $\rho_k^+ = \rho_k^-$ for all classes with one base loss, 
ours with $\L_{\RM}$ can assign class-aware $\rho_k^+$ and $\rho_k^-$ for distinct classes with two different base losses for each task. 
As $\rho_k^+ = {1}/{\Delta_k^+}$, we can adjust the margins $\Delta_k^+$ and $\Delta_k^+$ to make $\rho_k^+ + \rho_k^-$ smaller under the constant constraint of $\Delta_k^+ + \Delta_k^-$, leading to 
better guarantees.

\textbf{Choice of Optimal Margins.} Then, how to choose the optimal margins is critical.
Here we follow the idea in \citet{cao2019marginloss}, which proves that the optimal margins for binary classification are $\Delta^+ = \frac{\lambda}{{(n_1)}^{1/4}}$ and $\Delta^-=\frac{\lambda}{{(n_2)}^{1/4}}$, and extend it to multi-class classification with $\Delta_k=\frac{\lambda}{(n_k)^{1/4}}$ for each class $k$. For our margin choices, it becomes more complicated, including two main steps.

Firstly, in a two-label MLL task, we can derive the optimal margins for each class as $\widetilde{\Delta}_1 = \frac{\lambda}{{|\sD_1^+|}^{1/4}},\widetilde{\Delta}_2 = \frac{\lambda}{{|\sD_2^+|}^{1/4}}$ (see Proposition 1 in Appendix B.3).
Similarly to~\citet{cao2019marginloss}, we extend the results to an MLL task with more than two labels, getting $\widetilde{\Delta}_k = \frac{\lambda}{{|\sD_k^+|}^{1/4}}$.
This implies that distinct classes in a multi-label dataset should be given different margins associated with the number of positive instances. 

Secondly, for only one-label MLL task, indeed, it can be seen as a binary classification from positive and negative instances. Similarly to the above analysis, we analyze the optimal margins, obtaining $\Delta_k^+ = \Delta_k^-$ (see Proposition 1. (2) in Appendix B.3).
Intuitively, this result is somewhat surprising because it assigns equal margins to positive and negative instances for one label. 
This may be puzzling as it treats positive and negative instances equally.
However, we reweight the loss for positive and negative instances as the second equation in Eq.~\eqref{eq:emp_ldam}, implying that the positive and negative instances have been ``balanced". Hence, it makes sense to assign equal margins to both positive and negative instances.

Finally, we can get $\Delta_k^+ = \Delta_k^- = \frac{\lambda}{|\sD_k^+|^{1/4}}$ with a constant $\lambda$ as a hyper-parameter to be tuned in experiments.

\noindent\textbf{For RLDAM-based algorithm with WRU in MLCL.} 
Based on the above theory result in batch MLL, 
we can obtain the following learning guarantee of the RLDAM-based algorithm with WRU in MLCL, where the techniques mainly follow recent theory work~\citep{shi2024unified} in domain-incremental continual learning and the work~\citep{mansour2009domain} in domain adaptation.

\begin{restatable}[\textbf{Learning guarantee of RLDAM-based algorithm with WRU in MLCL, full proof in Appendix C}]
    {theorem}{LearningGuaranteeMLCL}
    \label{thm:learning_guarantee_mlcl}
    Suppose {\rm Assumption~\ref{assump_repetition} and \ref{assump_common}} hold. 
    Let $n^i$ and $\tilde{n}^i$ denote the number of samples in task $\T^i$ and from previous tasks in the memory buffer. According to class-incremental setup, let $f^{t,i}$ denote the outputs of function $f$ for a specific task $\T^i$ when learning task $\T^t$. Let $F^{t-1}$ be the model learned after $t-1$ tasks, and $F^{t-1,i}$ denote the outputs of model $F^{t-1}$ for a specific task $\T^i$. Assume constants $\alpha^i + \beta^i + \gamma^i = 1$. 
    Then, with probability at least $1-\delta$, we have 
    \begin{align*}
            \sum_{i=1}^{t} R_{\sD^i}(f^{t,i}) \le \Bigg\{ \sum_{i=1}^{t-1} \gamma^i \widehat{R}_{\sD^i}(f^{t,i}) + \widehat{R}_{\sD^t}(f^{t,t}) \Bigg\} & \\
            + \Bigg\{ \sum_{i=1}^{t-1} \gamma^i \epsilon^i +  \epsilon^t + 2\sum_{i=1}^{t-1} \beta^i (\epsilon^i+\epsilon^t) \Bigg\} & \\
            + \Bigg\{ \sum_{i=1}^{t-1} \gamma^i \xi^i + \xi^t + B\sum_{i=1}^{t-1} \beta^i (\xi^i+\xi^t) \Bigg\} & \\
            + \sum_{i=1}^{t-1}(\alpha^i + \beta^i)R_{\sD^i}(F^{t-1,i}) & \\
            + \Bigg\{ \sum_{i=1}^{t-1} \alpha^i \psi^i + \sum_{i=1}^{t-1} (\beta^i) \psi^t \Bigg\} 
            +  \sum_{i=1}^{t-1} \beta^i \mathtt{disc}(\sD^i,\sD^t), &
    \end{align*}
where $n^i = \tilde{n}^i$ when $i<t$, and
    \begin{align*}
            & \epsilon^i = \frac{4 \Lambda r^i }{\sqrt{n^i}} \bigg( \frac{1}{|\sK_i|} \sum_{k \in \sK_i} \sqrt{\frac{1}{\tau_k^i}}(\rho_k^{i+} + \rho_k^{i-}) \bigg), \\
             & \xi^i=6 B \sum_{i=1}^{t-1} \gamma^i \sqrt{\frac{\log(\frac{6}{\delta})}{2{n}^i}} \left( \sqrt{\frac{1}{|\sK_i|} \sum_{k \in \sK_i} \frac{1}{\tau_k^i}} \right), \\
             & \psi^i = R_{\sD^i}(f^{t,i},F^{t-1,i}).
    \end{align*}
\end{restatable}

\begin{remark*} 
    From the above bound of right terms, we can get valuable insights into the continual learning process, demonstrating the effectiveness of our proposed RLDAM loss and WRU.
    \textbf{(1)} The second term that involves all $\epsilon$ is the bound of model complexity which contains the $\rho_k^+, \rho_k^-$ for all tasks. Similar to the batch learning scenario, we can select better margins to make the bound tighter, \textbf{which indicates the effectiveness of our proposed RLDAM loss}.
    \textbf{(2)} In both the second and third terms there are $n^i \cdot \tau_k^i$ in denominator. Recall the definition of $\tau_k^i$, we can get $n^i \cdot \tau_k^i = \min\{\sD_k^{i+},\sD_k^{i-}\}$. 
    With our proposed WRU, we can guarantee that $n^i \cdot \tau_k^i$ for class in memory is \textbf{stable and unbiased} relative to multi-task joint training.
    \textbf{(3)} The fourth term is the \textbf{expected risk} of $F^{t-1}$ on all previous tasks, which reflects how well the model learned in the previous step affects the performance of the current model (\textbf{forward transfer}).
    \textbf{(4)} The fifth term measures \textbf{the gap between the current and previous model} on each task, which suggests that model updates that deviate too much from the previous in each step of learning will result in a larger bound, i.e., forgetting. The replay approach we used can be viewed as a regularization that restricts the degree of variation in the model. By applying WRU, this item can be reduced.
    \textbf{(5)} The last term is a measure of the \textbf{discrepancy distance} (defined in Definition 3, Appendix C.1) between the current task and all past tasks. If $\sD^i = \sD^t$, this term will be zero, thus reducing forgetting.
\end{remark*}

\section{Related Work}
\textbf{Continual Learning.}
Continual learning methods are mainly divided into three branches~\citep{de2021survey}: replay-based approaches~\citep{chaudhry2019ER,rebuffi2017icarl}, regularization-based approaches~\citep{zenke2017si,farajtabar2020orthogonal}, and architecture-based approaches~\citep{fernando2017pathnet,mallya2018packnet}. Here we focus on replay-based approaches. 

The rehearsal~\citep{de2021survey} (or experience replay~\citep{chaudhry2019ER,wang2023comprehensive}) approach is a branch of replay-based methods, which explicitly retrain on a subset of stored samples while training on new tasks. 
Concerning rehearsal memory, many approaches have been explored to exploit how to design and utilize it. 
Considering different tasks, several memory-updating strategies are proposed~\citep{chaudhry2019ER,riemer2018ltl,rebuffi2017icarl}.
Among them, Reservoir Sampling~\citep{vitter1985reservoir} is a sampling method commonly used to design memory updating strategies~\cite{riemer2018ltl,chaudhry2019ER}.

Theoretical study on CL is challenging even in the multi-class continual learning community. 
Several studies~\citep{knoblauch2020clhard,evron2023separabledata,lin2023theory2} on continual learning demonstrate the challenges involved in conducting theoretical analysis for continual learning.
However, we strive to theoretically derive a generalization bound for MLCL on replay-based framework following the proof in \citet{shi2024unified}, and verify the effectiveness of our proposed method.

\noindent\textbf{Multi-Label Continual Learning.}
PRS~\citep{kim2020prs} is an earlier work to tackle the MLL problem under CL settings, in which a Partition Reservoir Sampling (PRS) is proposed to maintain a balanced knowledge of both classes. 
\citet{liang2022ocdm} proposes optimizing class distribution in memory (OCDM) for MLCL. OCDM formulates the memory update mechanism as an optimization problem to minimize the distribution distance between the whole dataset and memory data and then updates the memory by solving this problem.
\citet{dong2023krt} proposes a knowledge restore and transfer (KRT) framework for Multi-Label Class-Incremental Learning, which includes a dynamic pseudo-label (DPL) module and an incremental cross-attention (ICA) module.

\vspace{-2mm}
\section{Experiments}
\label{sec:experiments}
In this section, we conduct experiments to illustrate the effectiveness of our method, which is summarized as follows:
(1) We conduct comparison experiments with other baselines to illustrate the superiority of our method.
(2) The memory size is influential to the replay-based approaches. We illustrate that our method consistently outperforms ER and is less sensitive to memory sizes.
(3) Ablation studies show the effect of each component proposed in our method.

Please see Appendix~E for more details and results.

\subsection{Experimental Setup}
\textbf{Benchmarks.} Following previous research in Multi-Label Continual Learning~\citep{kim2020prs,liang2022ocdm,dong2023krt}, we utilize three commonly used multi-label classification datasets: PASCAL VOC~\citep{everingham2015voc2012}, MSCOCO~\citep{lin2014coco} and NUS-WIDE~\citep{chua2009nus}. These datasets are then transformed into their continual versions as C-PASCAL-VOC, C-MSCOCO, and C-NUS-WIDE. More details are described in Appendix E.1.
For simplicity and space-saving, we use C-VOC, C-COCO, and C-NUS to represent three benchmarks.

\begin{table}

\scriptsize
\begin{center}
\begin{tabular}{llccc}
\toprule
              & Baseline & C-VOC & C-COCO & C-NUS \\
\midrule
\multirow{7}{*}{\rotatebox{90}{Macro-AUC}} & FT       & 53.62        & 69.46    & 66.91      \\
                     & FT-RLDAM    & 70.30        & 70.07    & 67.89      \\
                     & ER~\citep{chaudhry2019ER}       & 79.17        &    72.68      &      69.31      \\
                     & ER-RS~\citep{chaudhry2019ER}       & 79.18        & 73.07    & 70.14      \\
                     & PRS~\citep{kim2020prs}      & 76.96        & 70.83    & 69.64      \\
                     & OCDM~\citep{liang2022ocdm}     & 79.98        & 74.79    & 73.23      \\
                     & KRT~\citep{dong2023krt}    & 79.22   & 74.30  & 75.23      \\
                     & Ours     & \textbf{88.69}        & \textbf{77.93}    & \textbf{79.77}      \\
\midrule
\multirow{7}{*}{\rotatebox{90}{Forgetting}}   & FT       &     45.63         &     26.13     &      26.58 \\
                     & FT-RLDAM    &       23.24       &     30.15     &      30.30 \\
                     & ER~\citep{chaudhry2019ER}       & 3.44         &    17.56      &      8.64      \\
                     & ER-RS~\citep{chaudhry2019ER}          &      2.31        &    15.45      &         7.85   \\
                     & PRS~\citep{kim2020prs}         & 0.95       & \textbf{4.93}     & 8.83       \\
                     & OCDM~\citep{liang2022ocdm}       & \textbf{-0.19}        & 7.01     & \textbf{0.71}       \\
                     & KRT~\citep{dong2023krt}   & 3.93   & 1.50   & 5.60      \\
                     & Ours     & 2.05        & 8.22     & 8.72      \\
\bottomrule
\end{tabular}
\end{center}
\caption{Comparison results with other baselines on three benchmarks. The boldfaced items denote the best results.}
\label{table1}
\vspace{-3mm}
\end{table}

\noindent\textbf{Baselines.} We compare our method with several baselines, 
including plain ER (with a random sample strategy), ER-RS~\citep{chaudhry2019ER}, PRS~\citep{kim2020prs}, OCDM~\citep{liang2022ocdm} and KRT~\citep{dong2023krt}. ER and RS are commonly used in CL. PRS, OCDM and KRT are recent MLCL methods. Finetune performs sequential training among tasks without any CL techniques.

\subsection{Comparison Results}

We start by illustrating the imbalance statistics of an example of three tasks in C-MSCOCO, as shown in Fig.1, Appendix E. Task 1 exhibits varying sample sizes, with one class having around six thousand samples, while some others have only a few hundred. A similar phenomenon is observed in Task 2 and Task 3.

Regarding this kind of imbalance problem, we conduct experiments to demonstrate the effectiveness of our proposed method.
Table \ref{table1} shows that our method outperforms other baselines on all benchmarks. 
While our focus is on maximizing Macro-AUC using RLDAM loss and WRU, the forgetting metric is not our primary concern. However, our method still achieves agreeable forgetting performance. 
Comparing Finetune (FT) and Finetune with RLDAM loss (FT-RLDAM), we observe a significant improvement in Macro-AUC with RLDAM loss. This highlights our method's focus on enhancing overall and per-task performance. Moreover, with RLDAM loss, Finetune performs comparably to ER on C-MSCOCO and C-NUS-WIDE. 

\begin{figure}[tb]
    \begin{minipage}{0.49\textwidth}
        \centering
        \includegraphics[width=0.8\columnwidth]{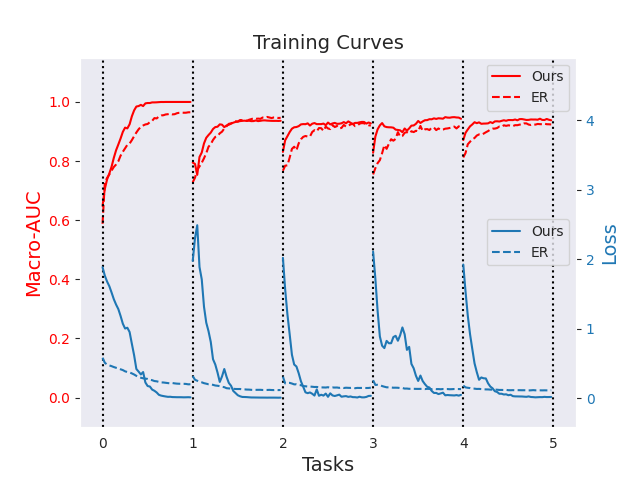}
        \caption{The comparison of training curves between our method and ER on C-PASCAL-VOC.}
        \label{fig2}
    \end{minipage}
    \hfill
    \begin{minipage}{0.49\textwidth}
        \centering
        \includegraphics[width=0.8\columnwidth]{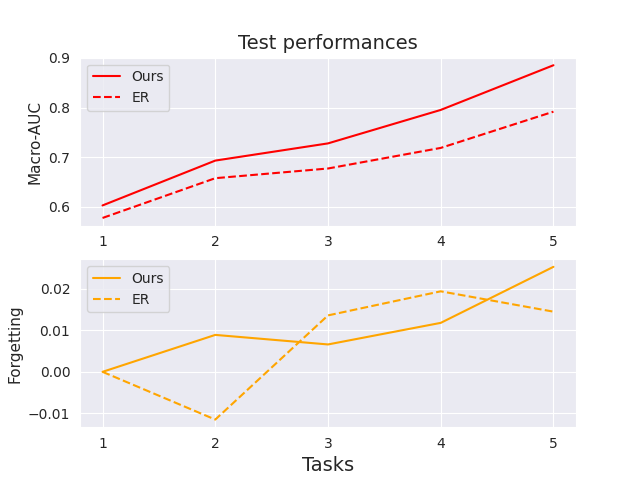}
        \caption{The comparison of overall test performances (test on all tasks after learning each single task) between our method and ER on C-PASCAL-VOC.}
         \label{fig:test}
    \end{minipage}
    \vspace{-5mm}
\end{figure}

Fig.\ref{fig2} presents the training curves on C-PASCAL-VOC. 
Our method consistently outperforms ER in Macro-AUC, except for the first task, and converges faster in terms of training loss. 
Fig.\ref{fig:test} demonstrates similar trends in test performances, with our method consistently achieving higher overall Macro-AUC compared to ER. 

\vspace{-1mm}
\subsection{Ablation Studies}

\begin{table}
\scriptsize
\centering
\begin{tabular}{lllccc}
\toprule
 Reweighted loss & Margin loss & WRU  &  C-VOC &  C-COCO & C-NUS \\
\midrule
$\times$ & $\times$ & $\times$ &  79.17 & 72.68 & 69.31\\
$\surd$ & $\times$ & $\times$ & 83.04 & 74.86 & 76.81  \\
$\surd$ & $\times$ & $\surd$ & 84.92 & 75.36 & 77.60  \\
$\times$ & $\surd$ & $\times$ & 78.80 & 73.95 & 76.03 \\
$\surd$ & $\surd$ & $\times$ & 86.29 & 77.61 & 78.59  \\
$\surd$ & $\surd$  & $\surd$ & \textbf{88.69} & \textbf{77.93} & \textbf{79.77}\\
\bottomrule
\end{tabular}
\caption{Ablation studies for each proposed component of our method. The $\surd$ denotes "with", and the $\times$ denotes "without". The first row without any component is actually ER with BCE loss. Note that ''$\times$" does not signify the absence of any replay strategy but rather the usage of a plain ER.}
\label{table:abla}
\end{table}

We conduct ablation experiments to validate the effect of each component that we propose. From the results shown in Table \ref{table:abla}, we can observe that the plainest ER just using BCE loss without any component we proposed underperforms others. When applying reweighted loss or margin loss, the Macro-AUC gains huge improvements. Furthermore, the combination of the two losses improves the performance further. The WRU retains the ratio of positive and negative instances, which tackles the imbalance in the memory and hence improves the overall Macro-AUC.

\vspace{-1mm}
\subsection{Effect of Memory Size}

\begin{table}[tb]
\scriptsize
\centering
\begin{tabular}{llccc}
\toprule
                    & Memory Size & C-VOC   & C-COCO & C-NUS \\
\midrule
\multirow{5}{*}{ER} & 200    &  67.36    &   63.36   &    63.94      \\
                    & 500      &   74.31    &  62.68    &      64.95   \\
                    & 1000     & 75.75 &   63.97   &      67.25  \\
                    & 1500     & 78.77 &   62.53   &      69.06   \\
                    & 2000     & 79.17 &   69.47   &     69.31     \\
\midrule
\multirow{5}{*}{Ours}      & 200      & \textbf{78.88} &   \textbf{75.78}   &     \textbf{76.08}     \\
                    & 500        & \textbf{81.86} &   \textbf{76.03}   &      \textbf{76.52}    \\
                    & 1000     & \textbf{85.70} &   \textbf{77.15}   &      \textbf{78.74}  \\
                    & 1500     & \textbf{87.89} &  \textbf{77.55}   &    \textbf{78.48}      \\
                    & 2000     & \textbf{88.49} &   \textbf{77.93}   &      \textbf{79.77}   \\
\bottomrule
\end{tabular}
\caption{The Macro-AUC of our method compared with ER under different memory sizes.}
\label{table:mem}
\vspace{-3mm}
\end{table}

We use the data replay approach as our CL technique. Consequently, we illustrate in this subsection that our proposed method performs well with different memory size setups. Specifically, we set five memory sizes: [200, 500, 1000, 1500, 2000]. The results are shown in Table \ref{table:mem}, from which we can see that our method consistently outperforms ER and is less sensitive to changes in memory size. That indicates our method is effective regardless of memory size.

\vspace{-1mm}
\section{Conclusion and Discussion}
In this paper, we focus on maximizing Macro-AUC for Multi-Label Continual Learning to tackle the imbalance problem, where the replay-based approach is considered.
To maximize Macro-AUC, we propose an RLDAM loss inspired by recent theoretical principled reweighted univariate loss and LDAM loss. 
Furthermore, we provide theoretical analyses of the generalization bound, supporting the superiority of our method. 
Then, based on the RLDAM loss and data replay framework, we propose the Weight Retain Updating (WRU) storing samples to maintain the numbers of positive and negative instances of each class into memory.
Finally, comprehensive experiments illustrate the effectiveness of our proposed method.

\section{Acknowledgments}

This work was supported by the NSFC (Nos. U23A20389, 62206159, 62176139), the Natural Science Foundation of Shandong Province (Nos. ZR2022QF117, ZR2021ZD15, ZR2024MF101), the Fundamental Research Funds of Shandong University, the Fundamental Research Funds for the Central Universities. G. Wu and H. Sun were also sponsored by the TaiShan Scholars Program (Nos. tsqn202306051, tsqn202312026).

\textbf{Ethical Statement.} Please see Appendix~F for limitations and impacts in detail. Our work can contribute to the MLCL community, while without any significant societal impacts.

\bibliography{aaai25}

\appendix
\onecolumn
\begin{center}
    \vbox{%
      \hsize\textwidth%
      \linewidth\hsize%
      \vskip 0.625in minus 0.125in%
      \centering%
      {\LARGE\bf Appendix for ``Towards Macro-AUC oriented Imbalanced Multi-Label Continual Learning" \par}%
      \vskip 0.1in plus 0.5fil minus 0.05in%
      \vspace{+0.3cm}
      {\Large{\textbf{Anonymous submission}}}%
      \vskip .2em plus 0.25fil%
      {\normalsize       }%
      \vskip .5em plus 2fil%
    }%
\end{center}
\vspace{+1.0cm}

\section{Learning guarantee of RLDAM-based algorithm in batch MLL}
First, we analyze the learning guarantee of the RLDAM-based algorithm w.r.t. the Macro-AUC measure in batch MLL. 

\begin{restatable}[\textbf{Learning guarantee of RLDAM-based algorithm in batch MLL, full proof in Appendix~\ref{sec_app:proof_theorem1}}]
    {theorem}{LearningGuaranteeURM}
    \label{thm:learning_guarantee_rm}
    Assume the loss $\L_{\phi} = \L_{\RM}$, where $\L_{\RM}$ is defined in Eq.~3, Section~3.1. Besides, {\rm Assumption~1 (Section~2) and Assumption~2 (Section~4)} holds. 
    Then, for any $\delta > 0$, with probability at least $1 - \delta$ over the draw of an i.i.d. sample $\sD$ of size $n$, the following generalization bound holds for any $f \in \mathcal{F}$: 
    \begin{align*}
        R^{0/1}(f)  \leq R^{\RM} (f) \leq  &  \widehat{R}_{\sD}^{\RM} (f) + \frac{4 \Lambda r 
        }{\sqrt{n}} \bigg( \frac{1}{K} \sum_{k=1}^{K} \sqrt{\frac{1}{\tau_k}}(\rho_k^+ + \rho_k^-) \bigg) + 6 B \sqrt{\frac{\log(\frac{2}{\delta})}{2n}} \left ( \sqrt{\frac{1}{K} \sum_{k=1}^K \frac{1}{\tau_k}} \right ) .
    \end{align*}
\end{restatable}

This theorem indicates that the batch algorithm with $\L_{\RM}$ has an imbalance-margin-aware learning guarantee of $O \left( \frac{1}{\sqrt{n}} \left( \frac{1}{K} \sum_{k=1}^K \sqrt{\frac{1}{\tau_k}}(\rho_k^+ + \rho_k^-) \right) \right)$ w.r.t. Macro-AUC. 
Compared with the learning guarantee of the $\L_{\RU}$-based algorithm (Theorem~4 in \citet{wu2023towards}), with identical $\rho$ for all classes, ours with $\L_{\RM}$ assigns class-aware $\rho_k^+$ and $\rho_k^-$ for distinct classes, which can provide better learning guarantees.

\section{Proof of Theorem~\ref{thm:learning_guarantee_rm}}
\label{sec_app:main_proof}

For subsequent analyses, we first draw on Definition 3 (page 4, \citet{wu2023towards}), and Theorem 1 (page 4,\citet{wu2023towards}) and give the the base theorem of learning guarantee for Macro-AUC.
\begin{definition} [\textbf{The fractional Rademacher complexity of the loss space, Definition~3 in~\citet{wu2023towards}}]
    For each label $k \in [K]$, construct the dataset $\tD_k = \{ (\tvx_{ki}, \ty_{ki}) \}_{i=1}^{m_k} = \{ ((\tvx_{ki}^+, \tvx_{ki}^-), 1) \}_{i=1}^{m_k}$ based on the original dataset $\sD_k$, where $(\tvx_{ki}^+, \tvx_{ki}^-) \in \sD_k^+ \times \sD_k^-$, and let $\{ (I_{kj}, \omega_{kj})\}_{j \in [J_k]}$ be a fractional independent vertex cover of the dependence graph $G_k$ constructed over $\tD_k$ with $\sum_{j \in [J_k]} \omega_{kj} = \chi_{f} (G_k)$, where $\chi_{f} (G_k)$ is the fractional chromatic number of $G_k$. For the hypothesis space $\mathcal{F}$ and loss function $L: \mathcal{X} \times \mathcal{X} \times \mathcal{F}_k \rightarrow \sR_+$, the empirical fractional Rademacher complexity of the loss space is defined as
    \begin{align*}
        \widehat{\mathfrak{R}}_{\tD}^*(L \circ \mathcal{F}) = \frac{1}{K} \sum_{k=1}^K \eE_{\boldsymbol{\sigma}} \Bigg[ \frac{1}{m_k} \sum_{j \in [J_k]} \omega_{kj} \times \sup_{f \in \mathcal{F}} \left( \sum_{i \in I_{kj}} \sigma_{ki} L(\tilde{\vx}_{ki}^+, \tilde{\vx}_{ki}^-, f_k) \right) \Bigg] .
    \end{align*}
\end{definition}

\begin{restatable}[\textbf{The base theorem of Macro-AUC, Theorem~1 in~\citet{wu2023towards}}]
    {theorem}{BaseTheoremMacroAUC}
\label{thm:base_theorem_macroauc}
Assume the loss function $\L_{\phi}: \mathcal{X} \times \mathcal{X} \times \mathcal{F}_k \rightarrow \mathbb{R}_+$ is bounded by $M$. Then, for any $\delta > 0$, the following generalization bound holds with probability at least $1 - \delta$ over the draw of an i.i.d. sample $\sD$ of size $n$: 
    \begin{align*}
        \forall f \in \mathcal{F}, \ R^{\phi} (f)  \leq \widehat{R}_{\sD}^{\phi} (f) + 
        2 \widehat{\mathfrak{R}}_{\tD}^*(\L_{\phi} \circ \mathcal{F})  +  3 M \sqrt{ \frac{1}{2n} \log \left(\frac{2}{\delta} \right)} \left( \sqrt{\frac{1}{K} \sum_{k=1}^K \frac{1}{\tau_k}}  \right) \ .
    \end{align*}
\end{restatable}

\subsection{The Fractional Rademacher Complexity of the Hypothesis Space}
Technically, we introduce a new definition of fractional Rademacher complexity of the hypothesis space w.r.t. variant of reweighted univariate losses and then get its upper bound.
Additionally, we propose a novel concentration inequality and provide rigorous theoretical proof to support it.


\begin{definition}[\textbf{The fractional Rademacher complexity of the hypothesis space w.r.t. variant of reweighted univariate losses}]
\label{def:fractional_rade_comp}
Given a dataset $\sD$ (and its corresponding constructed dataset $\tD$), assume the loss function 
$\L_{\phi} = \L_{\RU}(\vx^+, \vx^-, f_k) =  \ell(f_k(\vx^+)) + \ell(-f_k(\vx^-))$
for each label $k \in [K]$, where $\ell(\cdot)$ is the base loss function. 
In addition, for the latter proof, we assume a loss function $\Tilde{L}_{\phi}=\Tilde{L}_{\RU}(\vx^+,\vx^-,f_k)=\rho^+(\sD_k)\ell(f_k(\vx^+))+\rho^-(\sD_k)\ell(-f_k(\vx^-))$, where the function {$\rho^+(\sD_k)$ ( and $\rho^-(\sD_k)$)} is the margin-based function dependent on $\sD_k$.
Then, define the empirical fractional Rademacher complexity of the hypothesis space $\mathcal{F}$ w.r.t. $\tD$ and $\Tilde{L}_{u}$ as follows:
    \begin{align*}
        \widehat{\mathfrak{R}}_{\tD}^{\Tilde{L}_{\RU}}
        (\mathcal{F}) = \frac{1}{K} \sum_{k=1}^K \eE_{\boldsymbol{\sigma}} \left[ \frac{1}{m_k} \sum_{j \in [J_k]} \omega_{kj} \sup_{f \in \mathcal{F}} \left( \sum_{i \in I_{kj}} (\sigma_{ki}^+ {\rho}^+ (\sD_k) f_k(\vx_{ki}^+) + \sigma_{ki}^- {\rho}^- (\sD_k) f_k(\vx_{ki}^-) )  \right) \right].
    \end{align*}
\end{definition}

According to (Lemma 6, page 24, \citep{wu2023towards}), we can get: 
\begin{lemma}[\textbf{The upper bound of fractional Rademacher complexity of kernel-based hypothesis space w.r.t. variant of reweighted univariate losses}]
\label{lem:app_frac_rade_kernel_hypothesis_univariate_general}
Suppose (1) and (2) in {\rm Assumption~2} hold and the loss function $\L_{\RU}(\vx^+, \vx^-, f_k) =  \ell(f_k(\vx^+)) + \ell(-f_k(\vx^-))$. Then, for the kernel-based hypothesis space (Eq.~(7), Section~4), its empirical fractional Rademacher complexity w.r.t. the dataset $\tD$ and loss function $\L_u$, can be bounded as bellow:
    \begin{align*}
        \widehat{\mathfrak{R}}_{\tD}^{\Tilde{L}_{\RU}}
        (\mathcal{F}) \leq \frac{Br}{\sqrt{n}} \bigg( \frac{1}{K} \sum_{k=1}^{K} \sqrt{\frac{1}{\tau_k}}(\rho^+(\sD_k) + \rho^-(\sD_k)) \bigg).
    \end{align*}
    
\end{lemma}




Then we can establish the relationship between the fractional Rademacher Complexity of loss space and kernel-based hypothesis space.
\begin{lemma}[\textbf{A new contraction inequality for the RLDAM loss $\L_{\RM}$}]
\label{lem:app_contraction_univariate}
    For a dataset $\sD$ (and its corresponding constructed dataset $\tD$), assume the loss function $\L_{\phi} = \L_{\RM}(\vx^+, \vx^-, f_k) = \ell(f_k(\vx^+)-\Delta_k^+) +  \ell(-f_k(\vx^-)-\Delta_k^-)$ for each $k \in [K]$, where the two base loss functions $\ell(t-\Delta_k^+)$ and $\ell(t-\Delta_k^-)$ is $\rho_k^+$-Lipschitz and $\rho_k^-$-Lipschitz where $\rho_k$ indicates its dependency on $\Delta_k$. Then, the following inequality holds
    \begin{align*}
        \widehat{\mathfrak{R}}_{\tD}^*
        (\L_{\phi} \circ \mathcal{F}) \leq  2 
        \widehat{\mathfrak{R}}_{\tD}^{\Tilde{L}_{\RU}}
        (\mathcal{F}).
    \end{align*}
\end{lemma}

\begin{proof}
Since $\widehat{\mathfrak{R}}_{\tD}^*(\L_{\phi} \circ \mathcal{F}) = \frac{1}{K} \sum_{k=1}^K \widehat{\mathfrak{R}}_{\tD_k}^*(\L_{\phi} \circ \mathcal{F}_k)$ 
and $\widehat{\mathfrak{R}}_{\tD}^{\Tilde{L}_{\RU}}(\mathcal{F})
= \frac{1}{K} \sum_{k=1}^K \widehat{\mathfrak{R}}_{\tD_k}^{\Tilde{L}_{\RU}}(\mathcal{F})$, 
we first prove $\widehat{\mathfrak{R}}_{\tD_k}^*(\L_{\phi} \circ \mathcal{F}_k) \leq 2 \widehat{\mathfrak{R}}_{\tD_k}^{\Tilde{L}_{\RU}}(\mathcal{F})$ and then can get the desired result.
    
    Here we prove the inequality $\widehat{\mathfrak{R}}_{\tD_k}^*(\L_{\phi} \circ \mathcal{F}_k) \leq 2 \widehat{\mathfrak{R}}_{\tD_k}^{\Tilde{L}_{\RU}}(\mathcal{F})$ following the idea in~\citet{mohri2018foundations} (Lemma~5.7, p.93), and we omit the index $k$ and the symbol $\tD_k$ or $\sD$ for notation simplicity in the following.
    
    First we fix a sample $(\tvx_1=(\vx_1^+, \vx_1^-),\dots,\tvx_{m}=(\vx_m^+, \vx_m^-))$, then by definition,
    \begin{align*}
        \widehat{\mathfrak{R}}^*(\L_{\phi} \circ f) & =  \eE_{\boldsymbol{\sigma}} \left[ \frac{1}{m} \sum_{j \in [J]} w_{j} \sup_{f \in \mathcal{F}} \left( \sum_{i \in I_{j}} \sigma_{i} \L_{\RM}(\vx_i^+, \vx_i^-, f) \right) \right] \\
        & = \frac{1}{m} \sum_{j \in [J]} w_{j} \eE_{\sigma_1,\dots,\sigma_{n_j-1}} \left[ \eE_{\sigma_{n_j}} \left[ \sup_{f \in \mathcal{F}} u_{n_j - 1} (f) + \sigma_{n_j} \L_{\RM}(\vx_{n_j}^+, \vx_{n_j}^-, f)  \right ]\right] , \\
        & \qquad \qquad \qquad \qquad \qquad \qquad \qquad \qquad (\text{denote~} n_j = |I_j| \text{~for simplicity})
    \end{align*}
    where $u_{n_j- 1} (f) = \sum_{i=1}^{n_j} \sigma_i \L_{\RM}(\vx_i^+, \vx_i^-, f)$. By the definition of the supremum, for any $\epsilon > 0$, there exists $f^1, f^2 \in \mathcal{F}$ such that
    \begin{align*}
        u_{n_j - 1} (f^1) + \L_{\RU}(\vx_{n_j}^+, \vx_{n_j}^-, f^1) \leq (1 - \epsilon) \left[ \sup_{f \in \mathcal{F}} u_{n_j - 1} (f) +  \L_{\RM}(\vx_{n_j}^+, \vx_{n_j}^-, f)  \right ] ,
    \end{align*}
    and 
    \begin{align*}
        u_{n_j - 1} (f^2) - \L_{\RU}(\vx_{n_j}^+, \vx_{n_j}^-, f^2) \leq (1 - \epsilon) \left[ \sup_{f \in \mathcal{F}} u_{n_j - 1} (f) -  \L_{\RM}(\vx_{n_j}^+, \vx_{n_j}^-, f)  \right ] .
    \end{align*}
    Thus, for any $\epsilon > 0$, by definition of $\eE_{\sigma_{n_j}}$,
    \begin{align*}
        & (1 - \epsilon) \eE_{\sigma_{n_j}} \left[ \sup_{f \in \mathcal{F}} u_{n_j - 1} (f) + \sigma_{n_j} \L_{\RM}(\vx_{n_j}^+, \vx_{n_j}^-, f)  \right ] \\
        & = (1 - \epsilon) \left[ \frac{1}{2} \sup_{f \in \mathcal{F}} u_{n_j - 1} (f) +  \L_{\RM}(\vx_{n_j}^+, \vx_{n_j}^-, f)  \right ] + \left[ \frac{1}{2} \sup_{f \in \mathcal{F}} u_{n_j - 1} (f) -  \L_{\RM}(\vx_{n_j}^+, \vx_{n_j}^-, f)  \right ] \\
        & \leq \frac{1}{2} \left[ u_{n_j - 1} (f^1) +  \L_{\RM}(\vx_{n_j}^+, \vx_{n_j}^-, f^1)) \right] + \frac{1}{2} \left[ u_{n_j - 1} (f^2) -  \L_{\RM}(\vx_{n_j}^+, \vx_{n_j}^-, f^2) \right]. \\
        & = \frac{1}{2} \left[ u_{n_j - 1} (f^1) + \ell(f^1(\vx_{n_j}^+) - - \Delta_{nj}^+)) + \ell(-f^1(\vx_{n_j}^-)- \Delta_{nj}^-)) \right] \\
        & \quad + \frac{1}{2} \left[ u_{n_j - 1} (f^2) -   \ell(f^2(\vx_{n_j}^+)- \Delta_{nj}^+)) -  \ell(-f^2(\vx_{n_j}^-)- \Delta_{nj}^-)) \right].
    \end{align*}
    Let $s^+ = \sgn(f^1(\vx_{n_j}^+) - f^2(\vx_{n_j}^+))$ and $s^- = \sgn(f^1(\vx_{n_j}^-) - f^2(\vx_{n_j}^-))$. Then, the previous inequality implies
    \begin{align*}
        & (1 - \epsilon) \eE_{\sigma_{n_j}} \left[ \sup_{f \in \mathcal{F}} u_{n_j - 1} (f) + \sigma_{n_j} \L_{\RM}(\vx_{n_j}^+, \vx_{n_j}^-, f)  \right ]  \\
        & \leq \frac{1}{2} \left[ u_{n_j-1}(f^1) + u_{n_j-1}(f^2) + s^+ \rho^+ \left(f^1(\vx_{nj}^+) - f^2(\vx_{nj}^+)\right) + s^- \rho^- \left(f^1(\vx_{nj}^-) - f^2(\vx_{nj}^-)\right) \right] \\
        &  \qquad \qquad \qquad \qquad \qquad \qquad \qquad \qquad \qquad \qquad \qquad \qquad \qquad \qquad \qquad \quad (\text{Lipschitz property}) \\
        & = \frac{1}{2} \left[ u_{n_j-1}(f^1)+ s^+ \rho^+ f^1(\vx_{nj}^+)+ s^-\rho^- f^1(\vx_{nj}^-) \right] + \\
        & \qquad \frac{1}{2}\left[ u_{n_j-1}(f^2)  - s^+ \rho^+f^2(\vx_{nj}^+)  - s^-\rho^-f^2(\vx_{nj}^-) \right] \qquad \qquad (\text{rearranging}) \\
        & \leq \frac{1}{2} \sup_{f\in\mathcal{F}} \left[ u_{n_j-1}(f) + s^+\rho^+ f(\vx_{nj}^+) + s^-\rho^- f(\vx_{nj}^-) \right] + \\
        & \qquad \frac{1}{2} \sup_{f\in\mathcal{F}} \left[ u_{n_j-1}(f) - s^+\rho^+ f(\vx_{nj}^+) - s^-\rho^- f(\vx_{nj}^-) \right] \qquad \qquad (\text{def of sup}) \\
        & = 2\eE_{\sigma_{nj}^{+},\sigma_{nj}^{-}}\big[ u_{n_j-1}(\Tilde{f}) + \sigma_{nj}^{+}\rho^+ f(\vx_{nj}^+) + \sigma_{nj}^{-}\rho^- f(\vx_{nj}^-) \big] . \qquad \qquad (\text{definition of~}\eE_{\sigma_{n_j}^+,\sigma_{n_j}^-}) \\
    \end{align*}
    Since the inequality holds for any $\epsilon > 0$, we have
    \begin{align*}
        & \eE_{\sigma_{n_j}} \left[ \sup_{f \in \mathcal{F}} u_{n_j - 1} (f) + \sigma_{n_j} \L_{\RM}(\vx_{n_j}^+, \vx_{n_j}^-, f)  \right ] \\
        & \leq 2 \eE_{\sigma_{n_j}^+,\sigma_{n_j}^-} \left[ u_{n_j - 1} (\Tilde{f}) + \sigma_{n_j}^+ \rho^+ f(\vx_{n_j}^+) + \sigma_{n_j}^- \rho^- f(\vx_{n_j}^-) \right].
    \end{align*}
    Proceeding in the same way for all other $\sigma_i$ ($i \in [I_j], i \neq n_j$) proves that 
    \begin{align*}
         & \eE_{\boldsymbol{\sigma}} \left[ \frac{1}{m} \sum_{j \in [J]} w_{j} \sup_{f \in \mathcal{F}} \left( \sum_{i \in I_{j}} \sigma_{i} \L_{\RM}(\vx_i^+, \vx_i^-, f) \right) \right] \\
         & \leq  \eE_{\boldsymbol{\sigma}} \left[ \frac{1}{m} \sum_{j \in [J]} w_{j} \sup_{f \in \mathcal{F}} 2 \left( \sum_{i \in I_{j}} (\sigma_{i}^+ \rho^+ f(\vx_{i}^+) + \sigma_{i}^- \rho^- f(\vx_{i}^-) )  \right) \right]. 
    \end{align*}
    By proceeding other $j \in [J]$, we can obtain the following 
    \begin{align*}
        \widehat{\mathfrak{R}}^*(\L_{\phi} \circ f) & = \frac{1}{m} \sum_{j \in [J]} w_{j} \eE_{\boldsymbol{\sigma}} \left[ \sup_{f \in \mathcal{F}} \left( \sum_{i \in I_{j}} \sigma_{i} \L_{\RM}(\vx_i^+, \vx_i^-, f) \right) \right] \\
        & \leq \frac{1}{m} \sum_{j \in [J]} w_{j} \eE_{\boldsymbol{\sigma}} \left[ \sup_{f \in \mathcal{F}} \left( \sum_{i \in I_{j}} 2 (\sigma_{i}^+ \rho^+ f(\vx_{i}^+) + \sigma_{i}^- \rho^- f(\vx_{i}^-) )  \right) \right] \\
        & = 2 \widehat{\mathfrak{R}}^{\Tilde{L}_{\RU}}(f) .
    \end{align*}


\end{proof}

\subsection{Proof of Theorem~\ref{thm:learning_guarantee_rm}}
\label{sec_app:proof_theorem1}
\LearningGuaranteeURM*
\begin{proof}
    Since the base loss $\ell(z)$ is bounded by $B$ and the loss ${L}_{\RM}(\vx^+, \vx^-, f_k) = \ell \left ( f_k(\vx^+)-\Delta_k^+ \right ) + \ell \left ( -f_k(\vx^-)-\Delta_k^- \right)$, the loss $\L_{\phi} = \L_{\RM}$ is bounded by $2 B$. Then, applying Theorem~\ref{thm:base_theorem_macroauc}, and combining Lemma~\ref{lem:app_frac_rade_kernel_hypothesis_univariate_general} and Lemma~\ref{lem:app_contraction_univariate}, we can obtain that for any $\delta > 0$, the following generalization bound holds with probability at least $1 - \delta$ over the draw of an i.i.d. sample $\sD$ of size $n$:
    \begin{align*}
        R^{\RM}(f) =  \eE_{\sD} \left[ \widehat{R}_{\sD}^{\RM} (f) \right] 
        & \leq \widehat{R}_{\sD}^{\RM} (f) + \frac{4 \Lambda r 
        }{\sqrt{n}} \bigg( \frac{1}{K} \sum_{k=1}^{K} \sqrt{\frac{1}{\tau_k}}(\rho_k^+ + \rho_k^-) \bigg) + \\
        & 6 B \sqrt{\frac{\log(\frac{2}{\delta})}{2n}} \left ( \sqrt{\frac{1}{K} \sum_{k=1}^K \frac{1}{\tau_k}} \right )
    \end{align*}
    Finally, we can get the desired result.
\end{proof}

\subsection{The Optimal Margins for Multi-Label Dataset}
\label{sec_app:optimal_margin}
Next, we present the derived conclusion and the proof process regarding the optimal margin.

\begin{proposition}[]
\label{propos:opt_margin}
    For two-label learning, where each label can get a binary-classification problem, with Rademacher complexity upper bound $\widehat{\mathfrak{R}}_{\tD}^{\Tilde{L}_{\RU}} \le {\Lambda r} \bigg( \frac{1}{K} \sum_{k=1}^{K} \sqrt{\frac{1}{|\sD_k^+|}} \rho_k \bigg)$.
    \begin{enumerate}[(1)]
        \item 
        \label{propos:propos:opt_margin_1}
        Assume $\widetilde{\Delta}_k = \frac{1}{\rho_k}$. Suppose some two-label classifier $f=\{f_1,f_2\} \in \mathcal{F}$ can achieve a total sum of margins  and $\widetilde{\Delta}_1 + \widetilde{\Delta}_2 = \beta$ where the $\widetilde{\Delta}_1,\widetilde{\Delta}_2 > 0$ are constants.
    Then a classifier $f^* \in \mathcal{F}$ exists with margins
    \begin{align*}
        \widetilde{\Delta}_1^* = \frac{\beta {|\sD_2^+|}^{1/4}}{{|\sD_1^+|}^{1/4} + {|\sD_2^+|}^{1/4}},
        \widetilde{\Delta}_2^* = \frac{\beta {|\sD_1^+|}^{1/4}}{{|\sD_1^+|}^{1/4} + {|\sD_2^+|}^{1/4}}
    \end{align*}
    which gets the optimal generalization bound with probability $1-\delta$ for Theorem~\ref{thm:learning_guarantee_rm}.
    \begin{align*}
        R^{\RM}(f)  
        & \leq \widehat{R}_{\sD}^{\RM} (f) + 
        \min\limits_{{\widetilde{\Delta}_1}+{\widetilde{\Delta}_2}=\beta}{2 \Lambda r } \bigg(
        \frac{1}{\widetilde{\Delta}_1} \sqrt{\frac{1}{|\sD_1^+|}} + \frac{1}{\widetilde{\Delta}_2} \sqrt{\frac{1}{|\sD_2^+|}} \bigg) +  \epsilon^1 + \epsilon^2,
    \end{align*}
    where we denote $6 B \sqrt{\frac{\log(\frac{2}{\delta})}{2n}} \left ( \sqrt{\frac{1}{K} \sum_{k=1}^K \frac{1}{\tau_k}} \right )$ as $\epsilon^k$ for simplicity. 
    The optimal margins are $\widetilde{\Delta}_1^* = \frac{\lambda}{|\sD_1^+|^{1/4}},\widetilde{\Delta}_2^* = \frac{\lambda}{|\sD_2^+|^{1/4}}$.
        \item
        \label{propos:propos:opt_margin_2}
        Suppose a one label classifier (binary classifier) $f=\{f_1\} \in \mathcal{F}$ can achieve a total sum of margins $\Delta^+_1 + \Delta^-_1 = \beta_1$, $\Delta^+_2 + \Delta^-_2 = \beta_2$ where the $\beta_1,\beta_2 > 0$ are constants.
    Then take one label such as label $1$ for consideration, the relationship of margins is derived as $\Delta_1^+ = \Delta_1^-$, which obtains the optimal generalization bound with probability $1-\delta$ for Theorem~\ref{thm:learning_guarantee_rm}.
    \begin{align*}
        R^{\RM}(f)  
        & \leq \widehat{R}_{\sD}^{\RM} (f) + 
        \min\limits_{{\Delta_1^+}+{\Delta_1^-}=\beta_1}{2 \Lambda r }
        \sqrt{\frac{1}{|\sD_1^+|}} \left(\frac{1}{\Delta_1^+} + \frac{1}{\Delta_1^-} \right) +  \epsilon^1.
    \end{align*}
    Similarly, the generalization bound on label 2 can be obtained.
    \end{enumerate}

\end{proposition}
\begin{proof}
    We first prove (\ref{propos:propos:opt_margin_1}) in Proposition~\ref{propos:opt_margin}. By employing the same analytical approach, we can deduce the inference of (\ref{propos:propos:opt_margin_2}).
    
    To get the optimal $\widetilde{\Delta}_1^*,\widetilde{\Delta}_2^*$, we are actually to solve the following minimization:
    \begin{align*}
        \min\limits_{{\widetilde{\Delta}_1}+{\widetilde{\Delta}_2}=\beta}{2 \Lambda r } \bigg(
        \frac{1}{\widetilde{\Delta}_1} \sqrt{\frac{1}{|\sD_1^+|}} + \frac{1}{\widetilde{\Delta}_2} \sqrt{\frac{1}{|\sD_2^+|}} \bigg).
    \end{align*}
    By substituting the ${\widetilde{\Delta}_2} = \beta - {\widetilde{\Delta}_1}$ into the above equation and setting the derivative to zero, we can obtain the following result.
   \begin{align*}
       \frac{1}{(\beta - {\widetilde{\Delta}_1)^2 \sqrt{|\sD_2^+|}}} - \frac{1}{({\widetilde{\Delta}_1)^2 \sqrt{|\sD_1^+|}}} = 0.
   \end{align*} 
   Solving it gives $\widetilde{\Delta}_1^*$ and then the $\widetilde{\Delta}_2^*$ also got.
   
   Then we adopt similar analyses to get (\ref{propos:propos:opt_margin_2}) in Proposition~\ref{propos:opt_margin}.
    
\end{proof}

\section{For Continual Learning Setting}
\label{sec:app_B}
In this part, we extend the bound for the Batch Learning setting to the Continual learning setting. 
Our proof follows the proof process in \citet{shi2024unified}. Because in this part, we only focus on the RLDAM loss, for simplicity and better readability, we omit the superscript "RM" in the expected risk.
We first write the expected risk with respect to our RLDAM loss as 
\begin{align}
    R(f) = \frac{1}{K} \sum_{k=1}^K \eE_{\vx_p \sim P_k^+, \vx_q \sim P_k^-} \left[ \L_{\RM}(\vx_p, \vx_q, f_k) \right],
\end{align}

\subsection{Derive the bound for discrepancy distance}
\begin{definition}[\textbf{The discrepancy distance between distribution of two tasks}]
Assume two distributions $P$ and $Q$ have their own distinct label spaces $Y^P, Y^Q$. Let $f \in \mathcal{F}$ be a function $f=\{f^P,f^Q\}$ (single head function) in a hypothesis space $\mathcal{F}$.  The discrepancy distance $\mathtt{disc}$ between two distributions $P$ and $Q$ is defined as:
    \begin{align}
        \mathtt{disc}(P,Q) = \sup\limits_{f,f' \in \mathcal{F}} |
        & R_{P}(f^P,{f'}^P) - R_{Q}(f^Q,{f'}^Q) |.
    \end{align}
\end{definition}

\begin{lemma}
For any two hypothesis $f,f' \in \mathcal{F}$ and any two distributions $P, Q$, we have
    \begin{align}
        |R_{P}(f^P,{f'}^P) - R_{Q}(f^Q,{f'}^Q)| \le \mathtt{disc}(P,Q).
    \end{align}

    \begin{proof}
        By definition, we have
        \begin{align}
             \mathtt{disc}(P,Q) = \sup\limits_{f,f' \in \mathcal{F}} \big| 
             & R_{P}(f^P,{f'}^P) - R_{Q}(f^Q,{f'}^Q) \big| \ge \big| R_{P}(f^P,{f'}^P) - R_{Q}(f^Q,{f'}^Q) \big|
        \end{align}
    \end{proof}
\end{lemma}

\begin{proposition}[\textbf{Bound the discrepancy with Rademacher complexity}]
\label{propo:dis_bound}
    Assume the base loss function in RLDAM loss is bounded by $B$. Given a multi-label dataset $\sD$ with size $n$ and construct a dataset $\tilde{\sD}$ for it according to the construction principle in Definition~\ref{def:fractional_rade_comp}.  Let $P$ be a true distribution for constructed dataset $\tilde{\sD}$ and let $\widehat{P}$ denote the corresponding empirical distribution for $\tilde{\sD}$. Then for any $\delta > 0$, with probability at least $1-\delta$, we have
    \begin{align}
        \mathtt{disc}(P,\widehat{P}) \le 8\widehat{\mathfrak{R}}_{\tD}^{\Tilde{L}_{\RM}}(\mathcal{F}) + 6 B^2 \sqrt{\frac{\log(\frac{2}{\delta})}{2n}} \left ( \sqrt{\frac{1}{K} \sum_{k=1}^K \frac{1}{\tau_k}} \right ).
    \end{align}
    
    \begin{proof}
    We first re-write the Theorem~\ref{thm:learning_guarantee_rm} as below
    \begin{align}
        R^{\RM} (f) - \widehat{R}_{\sD}^{\RM} (f) \leq 4\widehat{\mathfrak{R}}^{{L}_{\RM}}(\mathcal{F}) + 6 B \sqrt{\frac{\log(\frac{2}{\delta})}{2n}} \left ( \sqrt{\frac{1}{K} \sum_{k=1}^K \frac{1}{\tau_k}} \right ).
        \label{eq:thm1_v2}
    \end{align}
    By Eq.\eqref{eq:thm1_v2} applied to $\mathcal{L}_{\RM}$ ( following the proof of Proposition 2, page 3, \citet{mansour2009domain}), we can get
    \begin{align}
        \mathtt{disc}(P,\widehat{P}) \le 4\widehat{\mathfrak{R}}^{{L}_{\RM}}(\L_{\RM} \circ \mathcal{F}) + 6 B^2 \sqrt{\frac{\log(\frac{2}{\delta})}{2n}} \left ( \sqrt{\frac{1}{K} \sum_{k=1}^K \frac{1}{\tau_k}} \right ).
    \end{align}
    Then applying Lemma~\ref{lem:app_contraction_univariate}, we can get the desired result.
    \end{proof}
\end{proposition}

\begin{corollary}
\label{coro:disc_bound}
Let $\mathcal{F}$ be a hypothesis space. Assume the base loss function in RLDAM loss is bounded by $B$.$\mathcal{L}_{\RM}$ bounded by $B$. 
Let $P$ be a true distribution and $\widehat{P}$ the corresponding empirical distribution for a sampled dataset $\widetilde{\sD}_P$.
Let $Q$ be a true distribution and $\widehat{Q}$ the corresponding empirical distribution for a sampled dataset $\widetilde{\sD}_Q$. 
Then for any $\delta>0$, with probability at least $1-\delta$ over samples $\widetilde{\sD}_P$ of size $m$ and samples $\widetilde{\sD}_Q$ of size $n$, we have
\begin{align}
    \mathtt{disc}(P,Q)   \le  \  & 8\widehat{\mathfrak{R}}_{\tD^P}^{\Tilde{L}_{\RM}}(\mathcal{F}) + 6 B^2 \sqrt{\frac{\log(\frac{4}{\delta})}{2m}} \left ( \sqrt{\frac{1}{|\sK^P|} \sum_{k \in \sK^P} \frac{1}{\tau_k}} \right ) + \mathtt{disc}(\widehat{P},\widehat{Q}) + \\
    & 8\widehat{\mathfrak{R}}_{\tD^Q}^{\Tilde{L}_{\RM}}(\mathcal{F}) + 6 B^2 \sqrt{\frac{\log(\frac{4}{\delta})}{2n}} \left ( \sqrt{\frac{1}{|\sK^Q|} \sum_{k \in \sK^Q} \frac{1}{\tau_k}} \right )
\end{align}

\begin{proof}
    By the triangle inequality, we have
    \begin{align}
        \mathtt{disc}(P,Q) & \le  \mathtt{disc}(P,\widehat{P}) + \mathtt{disc}(\widehat{P},\widehat{Q})+\mathtt{disc}(\widehat{Q},Q).
    \end{align}
    Then apply Proposition~\ref{propo:dis_bound} to $\mathtt{disc}(P,\widehat{P})$ and $\mathtt{disc}(\widehat{Q},Q)$, we can get the desired result.
\end{proof}
\end{corollary}

\subsection{Bound for continual tasks}
\label{sec_app:proof_theorem_mlcl}

The proof presented in the following is primarily based on the proof introduced in \citet{shi2024unified}. \citet{shi2024unified} specifically concentrates on the domain incremental learning setting, where each task shares the same label space, and they establish a unified bound using VC-Dimension.

In contrast, our focus lies on the more challenging class incremental learning setting, where new classes continuously emerge. Consequently, we derive a more tighter generalization bound employing Rademacher Complexity.

\begin{lemma}[\textbf{ERM based generalization bound}]
Let $f \in \mathcal{F}$ be an arbitrary function in the hypothesis space. When task $t$ arrives, there are $n^t$ points from task $t$ and $\tilde{n}^i$ data points from each task $i < t$. According to Eq.~(6), Section~3.1 and Theorem~\ref{thm:learning_guarantee_rm}, with probability at least $1-\delta$, we have:
    \begin{align*}
        R_{\T^t}(f) \le & \widehat{R}_{\T^t}(f) + \\
        &  \frac{4 \Lambda r^t}{\sqrt{n^t}} \bigg( \frac{1}{|\sK^t|} \sum_{k \in \sK^t} \sqrt{\frac{1}{\tau_k^t}}({\rho_k^t}^+ + {\rho_k^t}^-) \bigg) + 6 B \sqrt{\frac{\log(\frac{2}{\delta})}{2n^t}} \left ( \sqrt{\frac{1}{|\sK^t|} \sum_{k\in \sK^t} \frac{1}{\tau_k^t}} \right) + \\
        & \sum_{i=1}^{t-1} \left[ \frac{4 \Lambda r^i 
        }{\sqrt{\tilde{n}^i}} \bigg( \frac{1}{|\sK^i|} \sum_{k \in \sK^i} \sqrt{\frac{1}{\tau_k^i}}({\rho_k^i}^+ + {\rho_k^i}^-) \bigg) + 6 B \sqrt{\frac{\log(\frac{2}{\delta})}{2\tilde{n}^i}} \left ( \sqrt{\frac{1}{|\sK^i|} \sum_{k\in \sK^i} \frac{1}{\tau_k^i}} \right) \right].
    \end{align*}

    \begin{proof}
    \begin{align}
        R_{\T^t}(f) = \eE \limits_{\sD^1,...,\sD^t} \left[ \sum_{i=1}^{t} \left( \widehat{R}_{\sD^i}(f) \right) \right]  = \sum_{i=1}^{t} \left( \eE \limits_{\sD^i} \left[ \widehat{R}_{\sD^i}(f) \right] \right)  = \sum_{i=1}^{t} \left(  R_{\sD^i}(f)  \right).
    \end{align}
    Then combined with Theorem~\ref{thm:learning_guarantee_rm}, this proof can be completed.
\end{proof}
\end{lemma}

\begin{lemma}[\textbf{Intra-task model-based bound}]
\label{lemma:c2}
Let $f \in \mathcal{F}$ be an arbitrary function in the hypothesis space $\mathcal{F}$, and $F^{t-1}$ be the model trained after task $t-1$. The task-specific risk $R_{\sD^i}(f)$ on a previous task $\T^i$ has an upper bound: 
    \begin{align*}
        R_{\sD^i}(f^{t,i}) \le R_{\sD^i}(f^{t,i},F^{t-1,i}) + R_{\sD^i}(F^{t-1,i})
    \end{align*}
    where $R_{\sD^i}(f^{t,i},F^{t-1,i}) \triangleq \frac{1}{|\sK^i|} \sum_{k \in \sK^i} \eE_{\vx_p \sim P_k^+, \vx_q \sim P_k^-} \left[ |\L_{\RM}(\vx_p, \vx_q, f_{i,k}) - \L_{\RM}(\vx_p, \vx_q, F^{t-1,i,k})| \right]$ is the expectation of the gap between two hypothesis $f^{t,i}$ and $F^{t-1,i}$.

    \begin{proof}
        \begin{align*}
            R_{\sD^i}(f^{t,i}) & = R_{\sD^i}(f) - R_{\sD^i}(F^{t-1,i}) + R_{\sD^i}(F^{t-1,i}) \\ 
            & \le |R_{\sD^i}(f^{t,i}) - R_{\sD^i}(F^{t-1,i})| + R_{\sD^i}(F^{t-1,i}) \\ 
            & = \frac{1}{|\sK^i|} \sum_{k \in K_i} \eE_{\vx_p \sim P_k^+, \vx_q \sim P_k^-} \big[ |\L_{\RM}(\vx_p, \vx_q, f_{i,k}) - \L_{\RM}(\vx_p, \vx_q, F^{t-1,k})| \big] + R_{\sD^i}(F^{t-1,i}) \\
            & = R_{\sD^i}(f^{t,i},F^{t-1,i}) + R_{\sD^i}(F^{t-1,i}).
        \end{align*}
    \end{proof}
\end{lemma}

\begin{lemma}[\textbf{Cross-task model-based bound}]
\label{lemma:c3}
    Let $f \in \mathcal{F}$ be an arbitrary function in the hypothesis space $\mathcal{F}$, and $F^{t-1}$ be the model trained after task $t-1$. The task-specific risk $R_{\sD^i}(f)$ on a previous task $\T^i$ has an upper bound: 
    \begin{align}
        R_{\sD^i}(f^{t,i}) \le R_{\sD^t}(f^t,F^{t-1,t}) + \mathtt{disc}(P^i,P^t) +R_{\sD^i}(F^{t-1,i})
    \end{align}
    
    \begin{proof}
        \begin{align*}
            R_{\sD^i}(f^{t,i}) & \le R_{\sD^i}(f^{t,i},F^{t-1,i}) + R_{\sD^i}(F^{t-1,i}) \\
            & = R_{\sD^i}(f^{t,i},F^{t-1,i}) + R_{\sD^i}(F^{t-1,i}) - R_{\sD^t}(f^t,F^{t-1,t}) + R_{\sD^t}(f^t,F^{t-1,t}) \\
            & \le R_{\sD^t}(f^t,F^{t-1,t}) + |R_{\sD^i}(f^{t,i},F^{t-1,i}) - R_{\sD^t}(f^t,F^{t-1,t})| + R_{\sD^i}(F^{t-1,i}) \\
            & \le R_{\sD^t}(f^t,F^{t-1,t}) + \mathtt{disc}(P^i,P^t) + R_{\sD^i}(F^{t-1,i})
        \end{align*}
    \end{proof}
\end{lemma}

\begin{restatable}[\textbf{Learning guarantee of RLDAM-based algorithm with WRU in MLCL}]
    {theorem}{LearningGuaranteeMLCL}
    \label{thm:learning_guarantee_mlcl}
    Let $f \in \mathcal{F}$ be a function in a hypothesis space $\mathcal{F}$. Let $n^i$ and $\tilde{n}^i$ denote the number of samples in task $\T^i$ and samples from previous tasks in the memory buffer. Assume the base loss function in RLDAM loss is bounded by $B$. Besides, {\rm Assumption~1 and 2} holds.  According to class-incremental setup, let $f^{t,i}$ denote the outputs of function $f$ for a specific task $\T^i$ when learning task $\T^t$. Let $F^{t-1}$ be the model learned after $t-1$ tasks, and let $F^{t-1,i}$ denote the outputs of model $F^{t-1}$ for a specific task $\T^i$. Assume constants $\alpha^i + \beta^i + \gamma^i = 1$. Then,
    with probability at least $1-\delta$, we have 
    
    \begin{align*}
            \sum_{i=1}^{t} R_{\sD^i}(f^{t,i}) & \le \Bigg\{ \sum_{i=1}^{t-1} \gamma^i \widehat{R}_{\sD^i}(f^{t,i}) + \widehat{R}_{\sD^t}(f^{t,t}) \Bigg\} 
            + \Bigg\{ \sum_{i=1}^{t-1} \gamma^i \epsilon^i +  \epsilon^t + 2\sum_{i=1}^{t-1} \beta^i (\epsilon^i+\epsilon^t) \Bigg\}  \\
            & + \Bigg\{ \sum_{i=1}^{t-1} \gamma^i \xi^i + \xi^t + B\sum_{i=1}^{t-1} \beta^i (\xi^i+\xi^t) \Bigg\} 
            + \sum_{i=1}^{t-1}(\alpha^i + \beta^i)R_{\sD^i}(F^{t-1,i}) \\
            & + \Bigg\{ \sum_{i=1}^{t-1} \alpha^i \psi^i + \sum_{i=1}^{t-1} (\beta^i) \psi^t \Bigg\} 
            +  \sum_{i=1}^{t-1} \beta^i \mathtt{disc}(\sD^i,\sD^t), 
    \end{align*}
where $n^i = \tilde{n}^i$ when $i<t$, and
    \begin{align*}
            & \epsilon^i = \frac{4 \Lambda r^i }{\sqrt{n^i}} \bigg( \frac{1}{|\sK_i|} \sum_{k \in \sK_i} \sqrt{\frac{1}{\tau_k^i}}({\rho_k^i}^+ + {\rho_k^i}^-) \bigg), \\
             & \xi^i=6 B \sum_{i=1}^{t-1} \gamma^i \sqrt{\frac{\log(\frac{6}{\delta})}{2{n}^i}} \left( \sqrt{\frac{1}{|\sK_i|} \sum_{k \in \sK_i} \frac{1}{\tau_k^i}} \right), \\
             & \psi^i = R_{\sD^i}(f^{t,i},F^{t-1,i}).
    \end{align*}
\end{restatable}
\begin{proof}
    By applying Lemma~\ref{lemma:c2} ,Corollary~\ref{coro:disc_bound} and Lemma~\ref{lemma:c3} to each of the past tasks, we can have
    \begin{align*}
        R_{\sD^i}(f^{t,i}) & = (\alpha^i + \beta^i + \gamma^i)R_{\sD^i}(f^{t,i}) \\
        & \le {\alpha^i} \left[ R_{\sD^i}(f^{t,i},F^{t-1,i}) + R_{\sD^i}(F^{t-1,i}) \right] + \\
        & \quad {\beta^i} \left[ R_{\sD^t}(f^t,F^{t-1,t}) + \mathtt{disc}(P^i,P^t) +R_{\sD^i}(F^{t-1,i}) \right] + {\gamma^i}R_{\sD^i}(f^{t,i})
    \end{align*}
    Then re-organizing the terms, we have
    
    \begin{align}
        \sum_{i=1}^{t} R_{\sD^i}(f^{t,i}) \le & \bigg\{ \sum_{i=1}^{t-1} \left( \gamma^i R_{\sD^i}(f^{t,i}) + \alpha^i R_{\sD^i}(f^{t,i},F^{t-1,i}) \right) \bigg\} + \nonumber \\
        & \bigg\{ R_{\sD^t}(f^t) + \sum_{i=1}^{t-1} (\beta^i) R_{\sD^t}(f^t,F^{t-1,t})  \bigg\} + \nonumber \\
        & \sum_{i=1}^{t-1} \beta^i \mathtt{disc}(P^i,P^t) + \sum_{i=1}^{t-1}(\alpha^i + \beta^i)R_{\sD^i}(F^{t-1,i}).
    \end{align}

    From the definition, we know that $R_{\sD^i}(f^{t,i},F^{t-1,i}) = |R_{\sD^i}(f^{t,i}) - R_{\sD^i}(F^{t-1,i})|$. As $R_{\sD^i}(f^{t,i}), R_{\sD^i}(F^{t-1,i})$ are both expected risk, so $R_{\sD^i}(f^{t,i}) > 0, R_{\sD^i}(F^{t-1,i}) > 0$. Besides, the $F^{t-1}$ is a learned model which is fixed, so the $R_{\sD^i}(F^{t-1,i})$ can be seen as a constant.
    So we can have
    \begin{align}
        R_{\sD^i}(f^{t,i},F^{t-1,i}) = & |R_{\sD^i}(f^{t,i}) - R_{\sD^i}(F^{t-1,i})| \nonumber \\
        \le & \bigg|\widehat{R}_{\sD^i} (f^{t,i}) + \frac{4 \Lambda r^i}{\sqrt{n^i}} \bigg( \frac{1}{|\sK^i|} \sum_{k \in \sK^i} \sqrt{\frac{1}{\tau_k^i}}({\rho_k^i}^+ + {\rho_k^i}^-) \bigg) + \nonumber \\
    & 6 B \sqrt{\frac{\log(\frac{2}{\delta})}{2 n^i}} \left ( \sqrt{\frac{1}{|\sK^i|} \sum_{k \in \sK^i} \frac{1}{\tau_k^i}} \right ) - R_{\sD^i}(F^{t-1,i}) \bigg|.
    \end{align}
    
    For simplicity, we write the second and the third term in Theorem~\ref{thm:learning_guarantee_rm} as 
    \begin{align}
        & \epsilon^i = \frac{4 \Lambda r^i }{\sqrt{{n}^i}} \bigg( \frac{1}{|\sK^i|} \sum_{k \in K_i} \sqrt{\frac{1}{\tau_k^i}}({\rho_k^i}^+ + {\rho_k^i}^-) \bigg), \\
        & \xi^i=6 B \sum_{i=1}^{t-1} \gamma^i \sqrt{\frac{\log(\frac{6}{\delta})}{2{n}^i}} \left( \sqrt{\frac{1}{|\sK^i|} \sum_{k\in K_i} \frac{1}{\tau_k^i}} \right)
    \end{align}

    Then apply Theorem~\ref{thm:learning_guarantee_rm}, we can get
    \begin{align}
        \sum_{i=1}^{t} R_{\sD^i}(f^{t,i}) \le & \bigg\{ \sum_{i=1}^{t-1} \left(\gamma^i R_{\sD^i}(f^{t,i}) + \alpha^i R_{\sD^i}(f^{t,i},F^{t-1,i}) 
        \right) \bigg\} + \nonumber \\
        & \bigg\{ R_{\sD^t}(f^t) + \sum_{i=1}^{t-1} (\beta^i) R_{\sD^t}(f^t,F^{t-1,t}) \bigg\} + \nonumber \\
        & \sum_{i=1}^{t-1} \beta^i \mathtt{disc}(P^i,P^t) + \sum_{i=1}^{t-1}(\alpha^i + \beta^i)R_{\sD^i}(F^{t-1,i}) \nonumber \\
        \le & \Bigg\{ \sum_{i=1}^{t-1} \gamma^i \widehat{R}_{\sD^i}(f^{t,i}) + \sum_{i=1}^{t-1} \gamma^i \epsilon^i +  \sum_{i=1}^{t-1} \gamma^i \xi^i  + \sum_{i=1}^{t-1} \alpha^i R_{\sD^i}(f^{t,i},F^{t-1,i}) \Bigg\} + \nonumber \\
        & \Bigg\{ \widehat{R}_{\sD^t}(f^t) +  \epsilon^t + \xi^t + \sum_{i=1}^{t-1} (\beta^i) R_{\sD^t}(f^t,F^{t-1,t}) \Bigg\} + \sum_{i=1}^{t-1}(\alpha^i + \beta^i)R_{\sD^i}(F^{t-1,i}) \nonumber \\
        & + \sum_{i=1}^{t-1} \beta^i \bigg( \mathtt{disc}(\sD^i,\sD^t) + 2(\epsilon^i+\epsilon^t) + B(\xi^i+\xi^t) \bigg)
    \end{align}

    Reorganize the items, we can get
    \begin{align}
        \sum_{i=1}^{t} R_{\sD^i}(f^{t,i}) \le & \Bigg\{ \sum_{i=1}^{t-1} \gamma^i \widehat{R}_{\sD^i}(f^{t,i}) + \widehat{R}_{\sD^t}(f^t) \Bigg\} + \Bigg\{ \sum_{i=1}^{t-1} \gamma^i \epsilon^i +  \epsilon^t + 2\sum_{i=1}^{t-1} \beta^i (\epsilon^i+\epsilon^t) \Bigg\} + \nonumber \\
        & \Bigg\{ \sum_{i=1}^{t-1} \gamma^i \xi^i + \xi^t + B\sum_{i=1}^{t-1} \beta^i (\xi^i+\xi^t) \Bigg\} + \nonumber \\
        & \Bigg\{ \sum_{i=1}^{t-1} \alpha^i R_{\sD^i}(f^{t,i},F^{t-1,i}) + \sum_{i=1}^{t-1} (\beta^i) R_{\sD^t}(f^t,F^{t-1,t}) \Bigg\} + \nonumber \\
        & \sum_{i=1}^{t-1} \beta^i \mathtt{disc}(\sD^i,\sD^t) + \sum_{i=1}^{t-1}(\alpha^i + \beta^i)R_{\sD^i}(F^{t-1,i})
    \end{align}
\end{proof}

\section{Additional Related Work}
The study of multi-label learning, where each instance is associated with multiple labels, has been a focal area in machine learning due to its applicability in diverse domains. A critical challenge in this area is the imbalance issue, which can significantly influence the performance of learning algorithms.

The foundational work by \citet{tsoumakas2007multi} provided an initial framework for understanding multi-label learning, categorizing the approaches into problem transformation methods and algorithm adaptation methods. This early work laid the groundwork for subsequent studies focusing on more specific challenges such as label imbalance.
More recently, \citet{zhang2013multireview} addressed label imbalance by introducing the concept of label-specific features, which help in improving classification performance for infrequent labels.

Regarding imbalance issue in multi-class problem, a theoretical work \citep{cao2019marginloss} proposed a label-distribution-aware margin (LDAM) loss, which replaces the conventional cross-entropy objective, offers a theoretically principled approach motivated by minimizing a margin-based generalization bound.

Recent advancements in multi-label learning have been marked by a deeper exploration into the theoretical underpinnings of commonly used metrics.
\citet{wu2023towards} delve into the generalization properties of Macro-AUC, which, despite its widespread application, lacks comprehensive theoretical understanding. This work identifies a crucial factor influencing generalization bounds: label-wise class imbalance, and proposed a new reweighted univariate loss. Empirical results across various datasets further corroborate their theoretical findings, suggesting that addressing label-wise imbalance is essential for improving Macro-AUC generalization.

\section{Experimental Details}
\label{sec_app:experiments}

\subsection{Benchmarks}
\label{sec:a_baseline}
Several studies have focused on curating multi-label datasets for continual learning.
\citet{shmelkov2017incremental} select 20 out of 80 classes to create two tasks from MSCOCO, each with 10 classes.
\citet{kim2020prs} creates four tasks from MSCOCO and six tasks from NUS-WIDE, and the tasks are mutually exclusive.
These approaches discard a considerable number of classes and a significant amount of data from the dataset to achieve task disjointness. As a result, each task ends up with a limited amount of available data compared to the original dataset. 

Different from previous works, we suppose that the dataset should be utilized as completely as possible for learning. So we allow repeated instances between different tasks, but they are presented with different labels, i.e., they only contain task-specific class labels.
Specifically, we create tasks by selecting all the data involving a subset of classes from the entire dataset and masking out irrelevant classes. 
The class selection for each task is randomly determined and remains fixed during training. 
Eventually, we curate continual versions for the above three datasets as C-PASCAL-VOC, C-MSCOCO, and C-NUS-WIDE. We considered 5 tasks for C-PASCAL-VOC, 8 and 9 for C-MSCOCO and C-NUS-WIDE.

\begin{figure}[h]
\centering
\includegraphics[width=0.65\columnwidth]{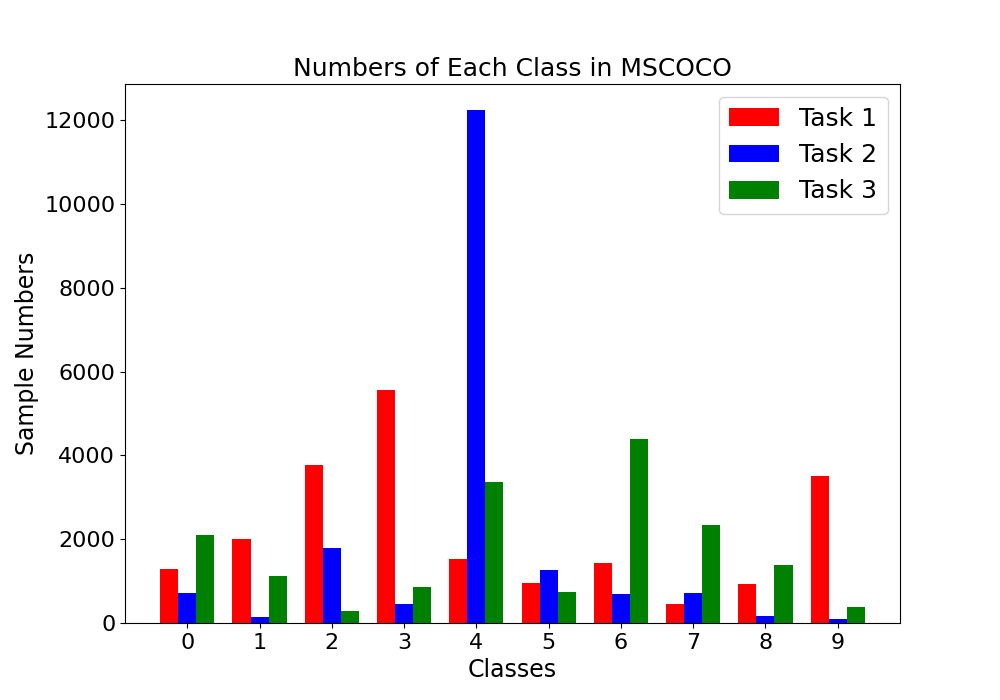}
\caption{Imbalance statistics of samples of each class in three tasks in C-MSCOCO.}
\label{fig1}
\end{figure}

\subsection{Implementation}
\label{sec:implementation}
We use ResNet~\citep{he2016resnet} as the base classifier\footnote{We do not use the pre-trained weights on ImageNet~\citep{deng2009imagenet} to avoid potential information leakage which may break the assumption that future information is unknown.}, specifically ResNet-34 for C-PASCAL-VOC and ResNet-50 for C-MSCOCO and C-NUS-WIDE. 
We use a fixed batch size of 128 for input data and memory data for C-PASCAL-VOC and a fixed batch size of 32 for C-MSCOCO and C-NUS-WIDE.
The SGD optimizer~\citep{robbins1951sgd} with a momentum of $0.9$ is employed, using a learning rate of $0.01$ and weight decay of $1e-5$. 
Regarding the PRS and OCDM originally tackling online CL tasks, we migrate the core methodology code of PRS and make a re-implementation based on the offline CL setting. 
As for OCDM, since it's still in a pre-printed version and has no released code, we do a re-implementation.
Due to the large size of C-MSCOCO and C-NUS-WIDE datasets and the time-consuming memory updates, training these benchmarks required more than 18 hours. Considering that and the limited stochastic processes involved in our approach, we report the results based on a single run. 
The hyper-parameter $\lambda$ in our method is set to $1.0,3.5,4$ for C-PASCAL-VOC, C-MSCOCO, and C-NUS-WIDE, respectively.
Our implementation is based on the released library Avalanche~\citep{lomonaco2021avalanche}.
The device we used to carry out our experiments was a Linux server running with 5 GeForce RTX 4090 (with memory of 24 GB), 256GB of server RAM, and 2TB of server storage space.

More specifically, at each iteration of training, we sample one batch $\sB^t$ from $\sD^t$ and another batch $\sB^\mathcal{M}$ from memory, respectively. The formulation of risk for a batch is similar to Eq.~(6), Section~3.1, and the SGD-based updating process is 
\begin{align}
    \Theta \gets \Theta - \eta \cdot \nabla \widehat{R}_{{\sB}^t}^{\RM}(f) - \eta \cdot \nabla \widehat{R}_{{\sB}^\mathcal{M}}^{\RM}(f),
\label{sgd_updat}
\end{align}

\textbf{Time Complexity Analyses.} The time complexity of the greedy algorithm for storing is $\mathcal{O} (|\sD^t| \times |\sM^t|)$. 
When the number of samples in a task is huge, the algorithm is time-consuming. Therefore, at each selecting step, we greedily select an instance from a random subset of the current task data. 
Assuming the size of the subset is $|{{\sD^t}^{\prime}}|$ ($|{{\sD^t}^{\prime}}| << |\sD^t|$), the time complexity is significantly reduced.

\subsection{Batch learning results}
\label{sec:batch_res}
Here we present the performance of our RLDAM loss on batch learning setup in Table~\ref{table:batch}. Note that when comes to batch learning, there is no WRU. The dataset we used here is the original multi-label dataset.
\begin{table}[h]
\centering
\begin{tabular}{lccc}
\toprule
Loss& VOC& COCO& NUS\\
\midrule
BCE&  74.58&   85.44&    84.77\\
RU&   83.03&  90.89&      92.68\\
RLDAM& \textbf{83.69}&   \textbf{90.93}&      \textbf{92.75}\\
\bottomrule
\end{tabular}
\caption{The Macro-AUC of our RLDAM loss on batch learning setup.}
\label{table:batch}
\end{table}

\subsection{Effect of hyper-parameter $\lambda$}
\begin{figure}[h]
\centering
\includegraphics[width=0.6\columnwidth]{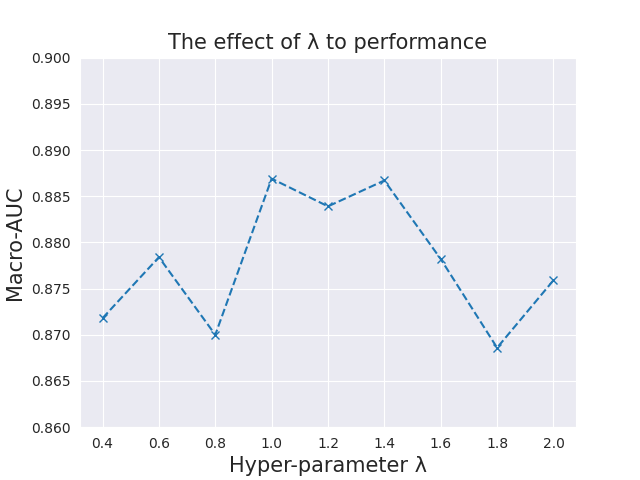}
\caption{The effect of hyper-parameter $\lambda$ to our RLDAM loss on C-PASCAL-VOC.}
\label{fig:lambda}
\end{figure}

  The result in Fig.~\ref{fig:lambda} is conducted on C-PASCAL-VOC and it shows that our method is not sensitive to the hyper-parameter $\lambda$. Under 11 different setups of $\lambda$ between 0 and 1, the Macro-AUC of our method is steady at around 0.89 and approaches an increasing and then decreasing trend. When $\lambda=0.7$, our method gets the best performance. Note that when $\lambda=0$, there is only reweighted loss, which indicates the effectiveness of the combination of reweighted loss and LDAM loss.

\subsection{Other than replay-based approach}
Other than replay-based approach, the loss function we proposed can actually be applied to all three approaches: replay-based, regularization-based, and architecture-based, as it is a multi-label classification loss. 
As discussed in \citet{knoblauch2020optimal,shi2024unified}, replay-based approaches often show superior performance and are easier to apply than other approaches, so we choose the replay-based approach as our continual learning framework.
\begin{table}[h]
\centering
\begin{tabular}{lccc}
\toprule
Loss& VOC& COCO\\
\midrule
BCE&  54.17&   67.28\\
RLDAM& \textbf{66.89}&   \textbf{70.78}\\
\bottomrule
\end{tabular}
\caption{The Macro-AUC of our RLDAM loss compared with BCE loss on EWC.}
\label{table:ewc}
\end{table}
As for other approaches, such as the regularization-based approach, we added a simple toy experiment to show the effectiveness of RLDAM loss on other approaches. We adopt a regularization-based approach EWC~\citep{kirkpatrick2017ewc} to conduct this comparison experiment between the commonly used binary cross-entropy loss and our RLDAM loss, and the results are shown in Table~\ref{table:ewc}.

\section{Ethical Statement}
\label{sec:limi}
Although we proposed to assign more proper margins in RLDAM loss to obtain a tighter generalization bound for both batch MLL and MLCL, there remain challenges to find the optimal margins for all tasks. 
Besides, we assume the model is a kernel function to conduct our theoretical analyses. Although we utilize deep neural networks (DNNs) in experiments, it can still provide valuable insights because recent theory~\citep{jacot2018ntk} has established the connection between over-parameterized DNNs and Neural Tangent Kernel (NTK)-based methods. To theoretically study the DNN using NTK is left as a future work.

This paper can contribute to the Continual Learning community and help theoretically understand the replay-based Macro-AUC-oriented MLCL algorithms, while without any significant societal impacts.

\end{document}